# Multi-encoder ConvNeXt network with smooth attentional feature fusion for multispectral semantic segmentation

Leo Thomas Ramos [a,*], Angel D. Sappa [a,b,*]

[a] Computer Vision Center, Universitat Autònoma de Barcelona, Barcelona, 08193, Spain
[b] ESPOL Polytechnic University, Guayaquil, 090112, Ecuador

## HIGHLIGHTS

- Introduction of MeCSAFNet, a dual-branch network tailored for multispectral semantic segmentation.
- Superior performance over baselines and state-of-the-art methods on the Potsdam dataset.
- Superior performance over baselines and state-of-the-art methods on the Five-Billion-Pixels dataset.
- Lightweight variants retain strong accuracy with reduced training requirements.
- Fast inference speeds across all versions enable real-time and large-scale deployment.

## ARTICLE INFO

*Keywords:*
Image segmentation
Semantic segmentation
Multispectral imaging
Land cover classification
Remote sensing
Computer vision
Convolutional neural networks
Earth observation
Satellite imagery
Urban mapping

## ABSTRACT

This work proposes MeCSAFNet, a multi-branch encoder–decoder architecture for land cover segmentation in multispectral imagery. The model separately processes visible and non-visible channels through dual ConvNeXt encoders, followed by individual decoders that reconstruct spatial information. A dedicated fusion decoder integrates intermediate features at multiple scales, combining fine spatial cues with high-level spectral representations. The feature fusion is further enhanced with CBAM attention, and the ASAU activation function contributes to stable and efficient optimization. The model is designed to process different spectral configurations, including a 4-channel(4c) input combining RGB and NIR bands, as well as a 6-channel(6c) input incorporating NDVI and NDWI indices. Experiments on the Five-Billion-Pixels(FBP) and Potsdam datasets demonstrate significant performance gains. On FBP, MeCSAFNet-base(6c) surpasses U-Net(4c) by +19.21%, U-Net(6c) by +14.72%, SegFormer(4c) by +19.62%, and SegFormer(6c) by +14.74% in mIoU. On Potsdam, MeCSAFNet-large(4c) improves over DeepLabV3+(4c) by +6.48%, DeepLabV3+(6c) by +5.85%, SegFormer(4c) by +9.11%, and SegFormer(6c) by +4.80% in mIoU. The model also achieves consistent gains over several recent state-of-the-art approaches. Moreover, compact variants of MeCSAFNet deliver notable performance with lower training time and reduced inference cost, supporting their deployment in resource-constrained environments. Model code is available at: https://github.com/Leo-Thomas/mecsafnet

## 1. Introduction

Remote Sensing (RS) is defined as the process of acquiring information from a distance [22], typically conducted via aircraft or satellites. This approach facilitates the analysis of various objects and phenomena without physical contact [1,22]. As a result, RS has become a highly valuable technology for obtaining data about the Earth [66]. Over the past decades, advancements in sensor design, satellite manufacturing, and data processing technologies have significantly improved the quality and availability of RS data [26]. These developments have led to higher resolutions across spatial, temporal, and spectral dimensions [66,83], enabling the collection of more detailed and comprehensive information. In turn, this progress has expanded the range of applications, establishing RS as a cornerstone for analyzing and addressing global phenomena [26].

One of the key advantages of RS, particularly through imaging, is its ability to capture information beyond the visible spectrum [47,57]. This type of data, referred to as Multispectral Imaging (MSI), enables the detailed analysis and recognition of objects and scenes that go beyond the capabilities of traditional RGB images [17]. By capturing information across multiple spectral bands [47], multispectral data provides a richer representation of the observed environment [56]. As a result, MSI has become a preferred approach in scenarios requiring precise detection, classification, and monitoring, making it an essential tool for numerous Earth observation applications [17].

* Corresponding authors at: Computer Vision Center, Universitat Autònoma de Barcelona, Barcelona, 08193, Spain.
*Email addresses:* ltramos@cvc.uab.cat (L.T. Ramos), asappa@cvc.uab.cat (A.D. Sappa).

Among the applications of RS and MSI, Land Cover Classification (LCC) has gained significant prominence due to its critical role in understanding the Earth's surface [81]. LCC involves the mapping and categorization of the various physical covers present on the Earth [57], such as vegetation, urban areas, water bodies, and soil [50]. It typically relies on segmentation techniques, where the goal is to delineate and classify different land cover types based on their spectral and spatial characteristics [23]. This makes LCC invaluable for diverse fields, including crop monitoring, urban planning, and environmental management [25,57].

Traditional approaches to LCC relied heavily on expert knowledge and hand-crafted feature engineering to interpret RS imagery [26,66]. However, understanding these images is a challenging task due to their inherent complexity [67,82], which arises from factors such as illumination variability, viewpoint changes, scale differences, noise, and their high spatial dimensionality [67,91]. These limitations often hindered the ability of such methods to deliver accurate and consistent classifications. In recent years, deep learning methods have emerged as the most prominent and widely adopted approaches in LCC [26,57]. These architectures are particularly effective in capturing abstract and semantic features that encode meaningful representations of the data [57,67], which they learn directly from large datasets. This capability allows them to handle the complex patterns and variations inherent in RS imagery [67].

Despite deep learning advancements, a common limitation of current approaches using MSI lies in their generalized treatment of spectral bands [57]. Many methods process all bands collectively, often without differentiating between the unique characteristics of the visible spectrum and other non-visible regions [48]. This lack of specialization ignores the complementary but distinct information these groups of bands offer [8], with the former capturing detailed spatial features and the latter providing richer spectral insights. Neglecting these inherent differences, such approaches risk losing critical details that are essential for precise analysis [36]. This can lead to reduced accuracy, especially in scenarios where leveraging the complementary nature of spectral information is crucial for distinguishing similar land cover types. In addition, effectively harnessing the potential of MSI requires robust strategies for fusing the features extracted from different spectral groups. Simple approaches, such as concatenation or averaging [13], often fail to model the complex relationships between spectral and spatial features. Consequently, exploring more effective fusion techniques has become a topic of significant interest [71], as these are essential for integrating the unique strengths of each group of bands and ensuring a more comprehensive analysis.

In response to these challenges, this work proposes Multi-encoder ConvNeXt Network with Smooth Attentional Feature Fusion (MeCSAFNet), an architecture for LCC specifically designed to better leverage the information provided by MSI for improved segmentation accuracy. The proposed architecture features a dual-branch encoder, where one branch processes information from the visible spectrum and the other focuses on non-visible spectral bands. For feature extraction in both branches, the ConvNeXt architecture is employed, a Convolutional Neural Network (CNN) known for its strong extraction capabilities, thanks to its design inspired by Transformers. For decoding and feature fusion, the architecture adopts a pyramid-based approach that progressively integrates multi-scale information from the encoded features. This process is further enhanced by attention modules during the fusion stage, enabling the model to emphasize the most relevant spectral and spatial details while suppressing irrelevant features. Additionally, the use of the ASAU activation function contributes to smoother and more stable optimization, especially in complex multispectral scenes.

Extensive experiments conducted on two multispectral datasets demonstrate that the proposed architecture effectively leverages MSI, delivering superior segmentation results compared to models that process all spectral channels jointly without specialized treatment. Experiments were carried out with two configurations: one combining RGB and NIR channels (4c), and another extending this input with NDVI and NDWI indices (6c). In both cases, MeCSAFNet consistently outperformed traditional baselines such as U-Net, DeepLabV3+, and SegFormer, as well as more recent specialized architectures, confirming its robustness and capacity to exploit the richness of multispectral information. The contributions of this work can be summarized as follows:

- We propose MeCSAFNet, a dual-branch architecture with a dedicated fusion decoder that explicitly separates the processing of visible and non-visible spectral inputs, enabling more effective exploitation of multispectral information.
- We demonstrate that the proposed approach consistently outperforms both traditional baselines and recent state-of-the-art methods on the Potsdam dataset.
- We show that MeCSAFNet achieves higher performance than both baselines and recent state-of-the-art architectures on the Five-Billion-Pixels dataset.
- We validate that lightweight variants of the proposed model maintain high segmentation accuracy while reducing training cost, making them suitable for scenarios with limited computational resources.
- We show that MeCSAFNet delivers fast inference across all variants, supporting its applicability in real-time or large-scale segmentation scenarios.

The remainder of this article is organized as follows. Section 2 reviews related work on multispectral segmentation and deep learning architectures. Section 3 details the proposed methodology, including the architecture design, datasets employed, and experimental setup. Section 4 presents the results and discussion, covering both quantitative and qualitative analyses. Section 5 outlines the main limitations of the current approach. Finally, Section 6 concludes the paper and outlines potential directions for future work.

This work is an extended version of the study presented at IEEE SoutheastCon 2025 [58]. It presents substantial new content and original contributions beyond the previous version, including an expanded and enhanced background, new architectural refinements, additional experiments conducted on a broader range of datasets and experimental scenarios, and a more comprehensive analysis of the results and the behavior of the proposed approach.

## 2. Related work

The field of semantic segmentation has seen significant advancements in recent years, particularly with the development of deep learning models tailored to handle multispectral RS imagery. While substantial progress has been made, challenges remain in effectively leveraging MSI beyond standard RGB inputs to enhance segmentation accuracy, especially in domains requiring detailed spatial and spectral understanding. As mentioned above, although MSI provides additional spectral information compared to RGB imagery, numerous approaches fail to fully exploit its potential. Many models simply stack the additional spectral bands alongside the RGB input without adapting their feature extraction strategies to account for the distinct characteristics of each spectral channel, often resulting in limited performance gains.

For instance, in [31], a U-Net model is applied to four-band (RGB-NIR) MSI data for classifying natural surface resources. The authors expand the network's input layer to accommodate the additional bands, comparing this approach to a standard RGB U-Net. Despite the increased spectral information, the four-channel approach achieved only a 0.51% improvement in overall accuracy over the RGB model. Similarly, Zheng and Chen [88] employ a U-Net for RGB-NIR segmentation, also expanding the input layer to accommodate the additional spectral channel. While the results slightly exceeded 80% in overall accuracy, questions arise about whether the architecture fully leverages the spectral information. This concern is amplified by the fact that the model was trained on over 100,000 patches but evaluated on only 840, which could indicate potential limitations in the architecture's ability to generalize effectively, despite the robust training setup. In [14], U-Net is also employed for RGB-NIR imagery, focusing on a specific use case in land mapping, with 768 images used as input. The methodology once again involves expanding the initial layer to process the four-channel data. While the maximum accuracy achieved was 84.8%, which is notable, it appears somewhat constrained given the limited number of training images. Moreover, the narrow scope of the study, tailored to a specific mapping scenario, raises questions about the generalizability of the approach to broader multispectral segmentation tasks.

Shifting focus from U-Net-based architectures, in [63], the Res-Seg-Net architecture is introduced. This network combines residual connections from ResNet with SegNet's nonlinear layers to create an autoencoder network. This approach utilizes six-band imagery; however, as with the previous methods, the bands were merely stacked at the input layer, resulting in maximum accuracies of 70%. CMPF-UNet [32] introduces ConvNeXt as the backbone of a U-shaped architecture, redesigning both the encoder and decoder with ConvNeXt bottleneck blocks and incorporating multiple pyramid-based fusion mechanisms to enhance multi-scale context modeling. While this approach significantly enriches feature representations within a single-stream pipeline, reaching 80.23% mIoU, it still processes multispectral inputs in a unified manner, without explicitly modeling spectral modality separation.

Other studies adopt multi-path architectures, but with a focus on feature characteristics rather than spectral modality separation. DPFE-AFF [28], for instance, proposes a dual-path framework that explicitly separates global contextual modeling from local texture extraction, drawing on ConvNeXt design strategies to strengthen the global branch. Although this design improves contextual awareness, the dual paths are defined by feature type (global versus local) rather than by visible and non-visible spectral information, and multispectral inputs remain implicitly coupled within the feature extraction process.

In response, modern approaches have explored more effective ways to utilize spectral information, with one widely studied method involving the use of separate branches to process the input data through independent feature extraction streams. In this context, selecting the appropriate backbones for feature extraction has become a key area of interest to maintain a proper balance between precision and efficiency. In [77], for example, a dual-encoder approach is proposed, with separate branches for processing RGB and NIR data. The encoders consist of two parallel, siamese branches, each built with convolutional and max-pooling layers organized across five stages. This method was tested and compared against U-Net models trained on RGB and RGB-NIR data, with the latter adapted by expanding its input layer. While the dual-encoder approach demonstrates superior performance compared to both U-Net variants, its overall accuracy of 78.1%, suggests that the additional spectral information was not fully leveraged, likely due to the simplicity of the encoder design. In [24], a different methodology is explored, employing two feature extraction branches based on U-Net to separately process RGB and NIR data. Although this approach outperforms the baseline U-Net by achieving accuracies above 70%, this performance is still underwhelming given the additional spectral information provided by the NIR channel, indicating that the reliance on duplicating the U-Net encoder may lack the robustness needed to fully capture and integrate the complexity of MSI, leaving significant room for improvement.

Another approach [4] employs two parallel ResNet101 networks to separately process RGB and NIR channels. ResNet101, known for its depth and capacity to capture complex features, is utilized in this approach with the expectation of improving segmentation accuracy by leveraging its hierarchical feature extraction capabilities. However, despite the use of such a large-scale architecture, the achieved accuracy only slightly exceeds 70%, raising concerns about whether the architecture effectively exploits the complementary information provided by the RGB and NIR channels, or if the parallel processing streams lack a robust mechanism for feature interaction and integration. Such methods, which duplicate the original encoders of baseline architectures and use ResNet as backbones, are common in the literature. However, they sometimes appear limited in their capacity to extract rich features, as reflected in relatively low metrics and inaccurate predictions.

To address these limitations, more robust feature extraction methods, such as Transformers, have been explored. In [49], for instance, a dual-branch architecture is introduced, leveraging ResNet and Swin Transformer (tiny version) networks to process RG-NIR images. Each branch independently processes the MSI input, and their extracted features are subsequently fused to produce the segmentation mask. This approach achieves accuracies slightly below 84%. Similarly, Gao et al. [12] adopt a comparable dual-branch approach for processing RG-NIR images. This method employs a ResNet34 network in parallel with a Swin Transformer, with their features combined using adaptive fusion modules. The approach achieves accuracies slightly above 86%, demonstrating an incremental improvement but still suggesting untapped potential for further optimization. In [35], a hybrid ConvNeXt–Transformer architecture is proposed for land-cover segmentation, combining convolutional feature extraction with Transformer-based global context modeling, together with positional encoding and spectral-aware augmentation to handle cross-sensor variability. While this design improves feature expressiveness reaching 88.10% and 78.40% on OA and mIoU respectively, it also increases architectural complexity and relies on additional training strategies beyond standard supervised learning.

As shown, applying more complex extractors brings a significant consideration: increasing the number of branches adds computational complexity and memory demands. Moreover, while Transformers are known for their outstanding performance in computer vision tasks, they inherently exhibit greater complexity than CNNs, which can lead to excessive training times and increased resource requirements, especially when processing RS images [10]. For these reasons, implementing such solutions becomes neither scalable nor efficient when multiple spectral groups are involved, requiring two or more branches. This becomes evident in the analyzed studies, as they seldom adopt fully Transformer-based approaches. Instead, their application introduces additional considerations, such as selecting a complementary branch architecture. As observed in [12], a ResNet34 is employed as the secondary branch, likely chosen for its lightweight architecture to counterbalance the computational demands of the Transformer. Similarly, Ma et al. [49] opts for the lighter variant of the Swin Transformer, an evident attempt to balance performance and resource constraints. These choices reflect the challenges of incorporating Transformers in multispectral segmentation, particularly when scalability and efficiency are priorities.

The analyzed works underscore key limitations in current approaches to leveraging MSI for semantic segmentation. Many methods rely on straightforward strategies, such as stacking additional spectral channels or duplicating encoder branches, which often remain insufficient to capture the full spatial and spectral complexity of multispectral data. As a result, performance gains are frequently marginal despite the availability of additional spectral information and, in some cases, large training datasets.

Recent advances have explored stronger feature extractors, including Transformer-based components, aiming to enhance contextual modeling capabilities. However, these approaches typically process multispectral inputs in a unified or feature-centric manner, or introduce substantial computational and architectural overhead. Consequently, they often require design compromises, such as simplified complementary branches, or restricted applicability to smaller-scale settings.

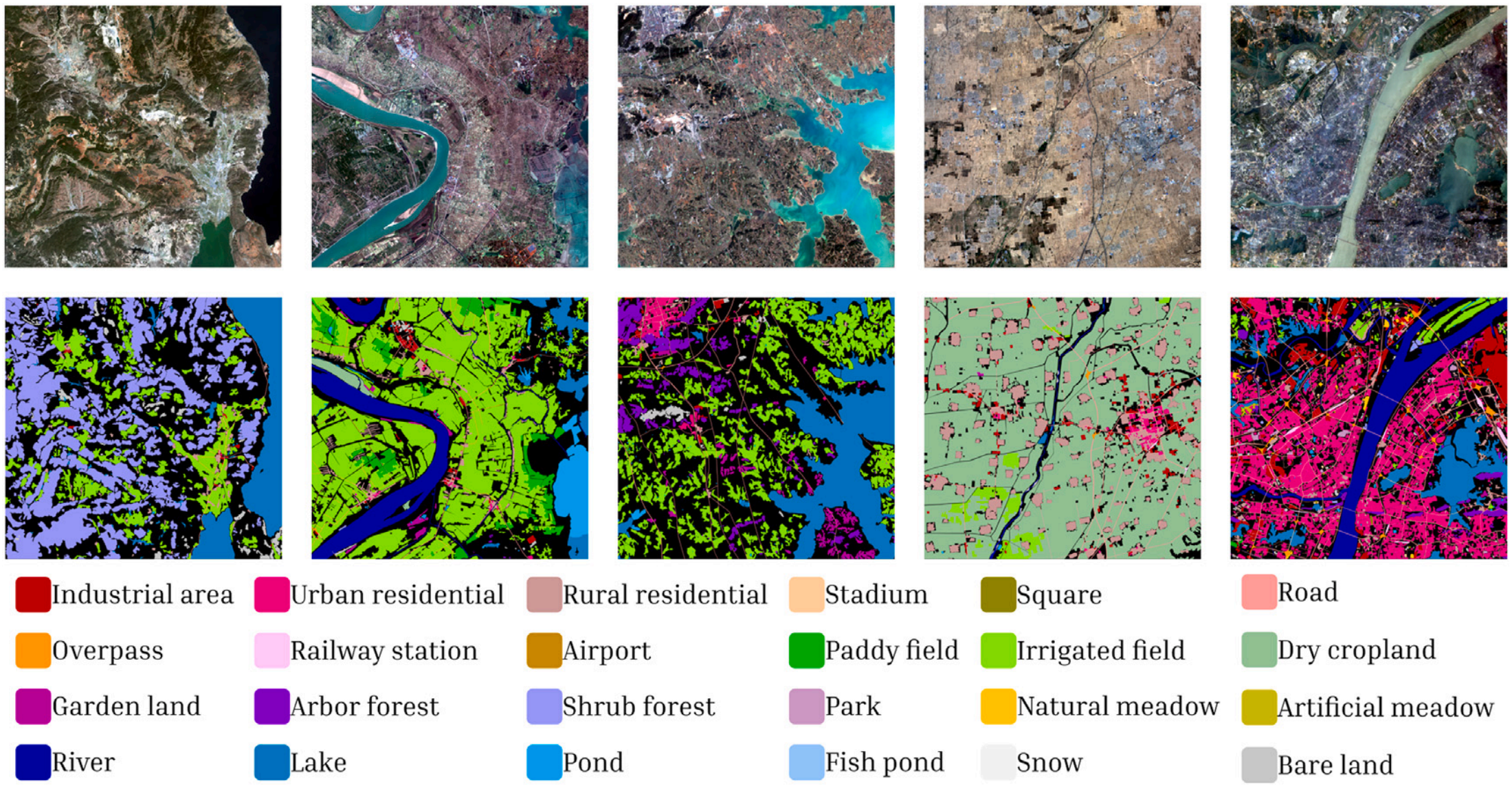

**Fig. 1.** Example images of the Five-Billion-Pixels dataset.

These observations highlight a gap in existing methodologies: the lack of architectures that explicitly disentangle visible and non-visible spectral information while maintaining strong feature extraction capacity and practical computational constraints. This gap motivates the proposed architecture, which adopts a dual-branch, modality-aware design to separately process visible and non-visible spectral components. The model employs ConvNeXt backbones together with an effective decoder structure, enabling the full exploitation of the spatial and spectral richness of MSI without imposing excessive resource demands compared to existing baseline models.

## 3. Methods

### *3.1. Dataset description*

To evaluate the proposed approach under diverse conditions, we have selected two publicly available multispectral datasets: Five-Billion-Pixels and Potsdam. These datasets pose distinct challenges that test different aspects of the model's design. The Five-Billion-Pixels dataset involves large-scale imagery with a high number of classes, demanding fine-grained discrimination and consistent classification over extensive and varied regions. In contrast, the Potsdam dataset emphasizes spatial accuracy due to its very high resolution and the presence of small urban structures. This setup enables us to assess the model's adaptability to both geometric detail and semantic complexity. Below, we describe each dataset.

#### *3.1.1. Five-Billion-Pixels*

The first dataset used in this work is Five-Billion-Pixels (FBP) [69], a relatively new dataset released in 2023 and built as an extension to the widely-known GID dataset [68]. This dataset increases the number of classes from 15 in GID to 24, along with an additional unlabeled class, providing a more challenging and extensive benchmark for the development of LCC models. The dataset is derived from Gaofen-2 satellite imagery, with a spatial resolution of 4 m per pixel. It includes over five billion labeled pixels across 150 satellite images, covering more than 50,000 square kilometers of various regions in China. The images have a resolution of $6800 \times 7200$ pixels and contain four channels: RGB-NIR. The dataset is available in both 16-bit and 8-bit spectral resolutions, with the images provided in TIFF format and the masks in PNG format. Examples from this dataset are shown in Fig. 1.

#### *3.1.2. ISPRS potsdam*

The second dataset used in this work is the Potsdam[1] dataset, created by the International Society for Photogrammetry and Remote Sensing (ISPRS). It consists of 38 large tiles, each with a resolution of $6000 \times 6000$ pixels, captured using airborne instruments over the city of Potsdam, Germany. The images have an 8-bit spectral resolution and are available in three band configurations: RGB, RG-NIR, and RGB-NIR. This study utilizes the RGB-NIR configuration. The dataset includes pixel-level annotations spanning six categories: impervious surfaces, buildings, low vegetation, trees, cars, and clutter/background. Additionally, the images are provided in TIFF format and are accompanied by digital surface models (DSM) with a ground sampling distance of 5 cm. Examples from this dataset are shown in Fig. 2.

### *3.2. Model design*

Overall, the proposed model is based on an encoder–decoder architecture, as depicted in Fig. 3. In the encoding stage, the model processes visible and non-visible data streams independently, leveraging ConvNeXt architectures to extract hierarchical features that capture both spatial and semantic information. These features are then passed to their respective decoders, which progressively reconstruct spatial details across multiple stages. In parallel, an intermediate fusion branch integrates features from both decoders at each stage to combine low-level spatial details with high-level information. This multi-branch design ensures the model effectively leverages the complementary nature of multispectral information, allowing for a more precise integration of fine-grained spatial details with meaningful spectral features, thereby improving the accuracy and robustness of semantic segmentation.

#### *3.2.1. ConvNeXt encoder*

For the feature extractor branches in our proposed model, the ConvNeXt [46] architecture was chosen. ConvNeXt is a CNN designed to achieve performance comparable to state-of-the-art Vision Transformers while retaining the advantages of traditional CNNs. To accomplish this, the authors began with a ResNet50 as the base architecture and implemented a series of structural modifications mainly inspired by the Swin Transformer.

[1] https://www.isprs.org/resources/datasets/benchmarks/UrbanSemLab/2d-sem-label-potsdam.aspx

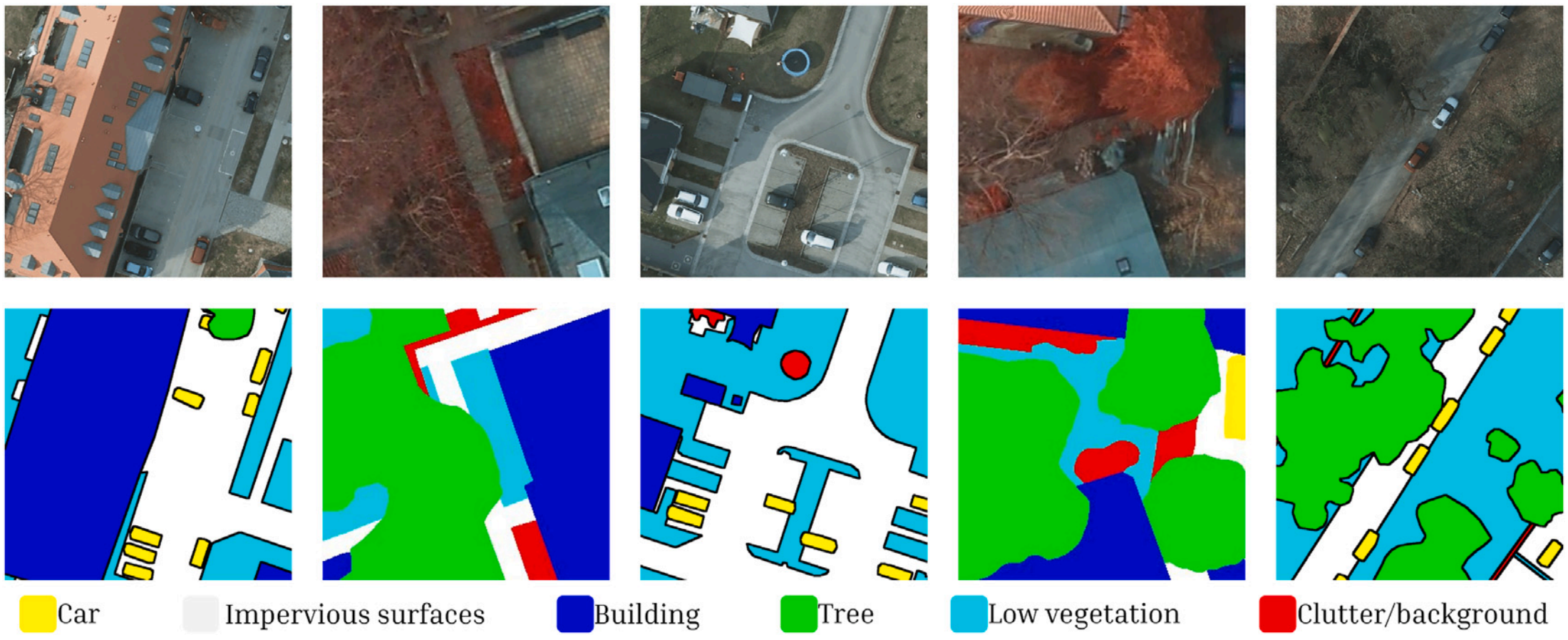

**Fig. 2.** Example images of the Potsdam dataset.

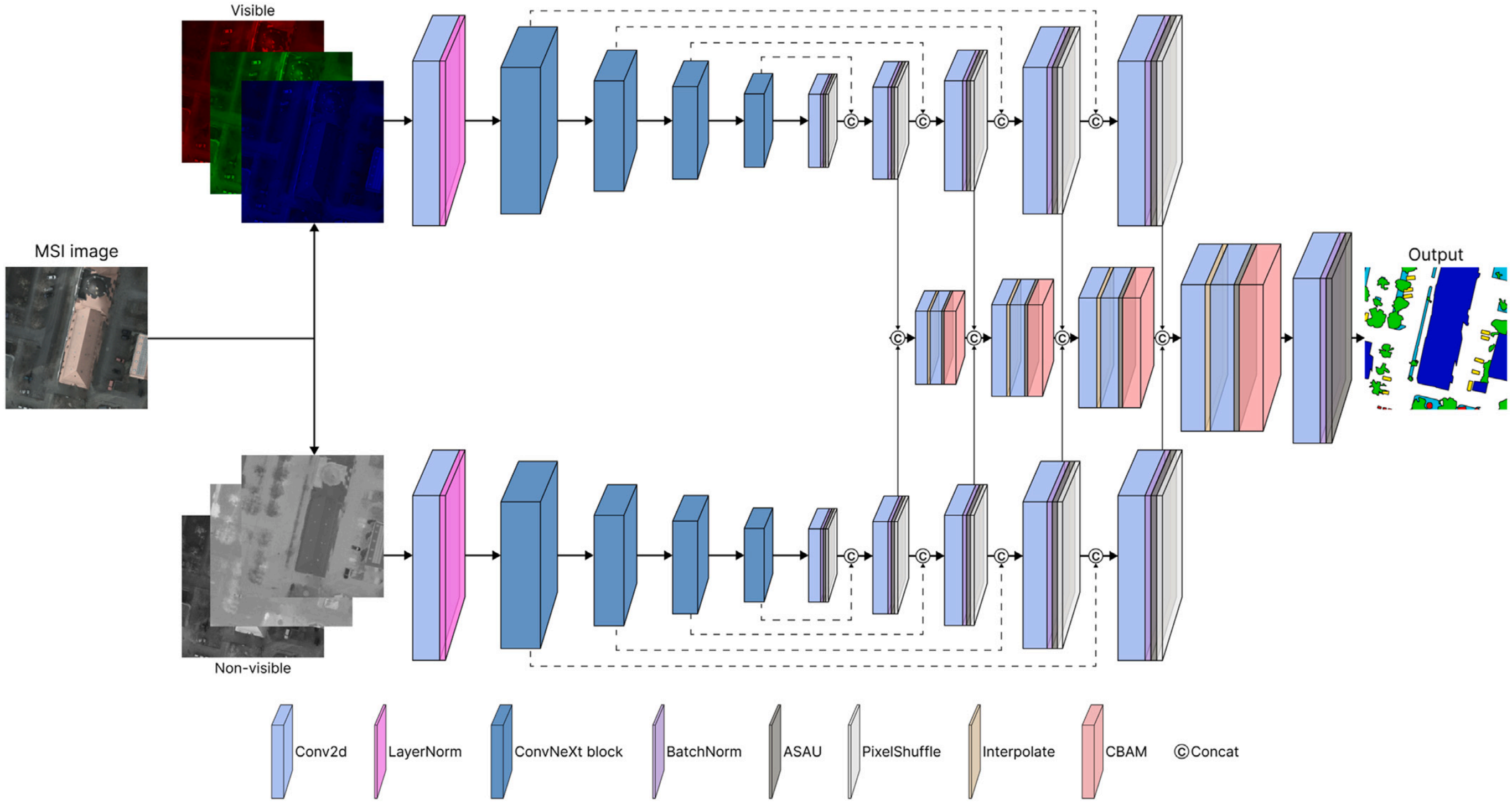

**Fig. 3.** Overview of the MeCSAFNet model proposed in this work.

To begin, the authors incorporate the multi-stage design present in the Swin Transformer into ConvNeXt. This design divides the architecture into stages, each operating on a different feature map resolution. This allows the model to progressively capture hierarchical representations, where lower stages focus on fine-grained details, and higher stages extract more abstract features. Another modification involves adopting grouped convolutions. This approach divides the convolutional kernels of a layer into separate groups. This allows the network to produce multiple channel outputs per layer, enabling wider architectures without increasing computational cost and facilitating the learning of varied and rich features. Furthermore, ConvNeXt incorporates the inverted bottleneck design, similar to that in the Swin Transformer. This approach allows the hidden dimension of the multi-layer perceptrons to be larger than the input dimension, enabling the model to capture more information in a larger latent space while simultaneously reducing FLOPs. The architecture also incorporates an increase in kernel size, shifting from the $3 \times 3$ kernels in ResNet to $7 \times 7$ kernels presented in the Swin Transformer. This change allows the model to capture a larger receptive field in each convolutional operation, enabling it to better represent spatial relationships and extract features across a broader context.

In addition to the aforementioned modifications, a series of minor changes were also implemented, including replacing the ReLU activation function with GELU, substituting batch normalization with layer normalization, decoupling subsampling layers, and reducing the number of normalization layers. These modifications resulted in the ConvNeXt architecture, whose performance testing on relevant benchmarks demonstrated superior results compared to the baseline ResNet and even Swin Transformers [2,11,46,61,62], while maintaining lower complexity than the latter. Fig. 4 presents a comparison between the blocks of ResNet, Swin Transformer, and ConvNeXt, illustrating the architectural modifications in the latter.

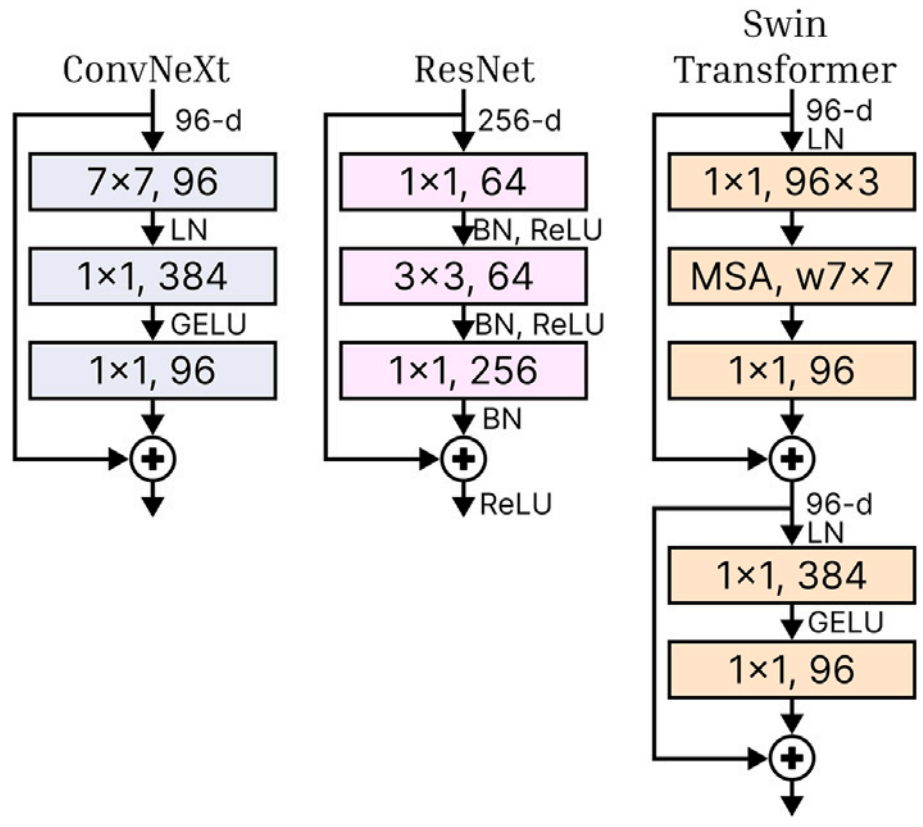

**Fig. 4.** Comparison between ResNet, Swin Transformer, and ConvNeXt blocks.

The complete ConvNeXt architecture consists of four stages, each containing a different number of ConvNeXt blocks, with the third stage having the highest count, as shown in Fig. 5. As the input progresses through the stages, features are extracted hierarchically, with finer details being captured in the later stages. At the end of the feature extraction process, the baseline ConvNeXt architecture incorporates a global average pooling layer followed by a fully convolutional layer as the classifier. The ConvNeXt architecture is available in four variants: Tiny, Small, Base, and Large, whose details are summarized in Table 1. In our implementation, these ConvNeXt variants define the corresponding MeCSAFNet versions: MeCSAFNet-tiny, MeCSAFNet-small, MeCSAFNet-base, and MeCSAFNet-large, each incorporating one of the four ConvNeXt models as encoders, respectively. Details of MeCSAFNet variants are shown in Table 2.

**Table 2**
Parameter count for MeCSAFNet variants. The dual-branch design increases the total number of parameters compared to single-branch models. Multiple variants are therefore defined to support different practical settings, from lightweight to higher-capacity configurations. As shown later, the lighter variants remain comparable to single-branch baselines in terms of training and inference efficiency (see Section 4).

| Model | Encoder | Params (M) |
|---|---|---|
| MeCSAFNet-tiny | ConvNeXt-tiny | 78.01 |
| MeCSAFNet-small | ConvNeXt-small | 121.28 |
| MeCSAFNet-base | ConvNeXt-base | 204.28 |
| MeCSAFNet-large | ConvNeXt-large | 435.17 |

The motivation for using ConvNeXt lies in its ability to strike a balance between performance and computational complexity. While CNNs such as ResNet perform well in various scenarios, LCC often requires a greater capacity to distinguish between diverse classes. In complex settings, CNN-based architectures have been shown to lag behind the performance of more advanced approaches like Transformers [16]. However, despite their superior accuracy, Transformers exhibit significantly higher computational complexity compared to traditional CNNs. This increased complexity poses practical limitations, particularly for multi-encoder paradigms, which require multiple backbone networks and would dramatically escalate computational and memory demands. Although efforts have been made to improve the efficiency of Transformers, processing multispectral remote sensing images remains computationally intensive, as highlighted by Fan et al. [10]. Given these considerations, our proposed model integrates ConvNeXt as its backbone, ensuring robust performance while addressing the challenges of computational efficiency required for multispectral LCC.

In the proposed approach, two parallel ConvNeXt networks are employed for feature extraction, with one branch processing visible information and the other handling non-visible spectral information. The architecture supports two input configurations: a 4-channel setup using RGB-NIR bands, and a 6-channel configuration that additionally incorporates the Normalized Difference Vegetation Index (NDVI) and the Normalized Difference Water Index (NDWI).

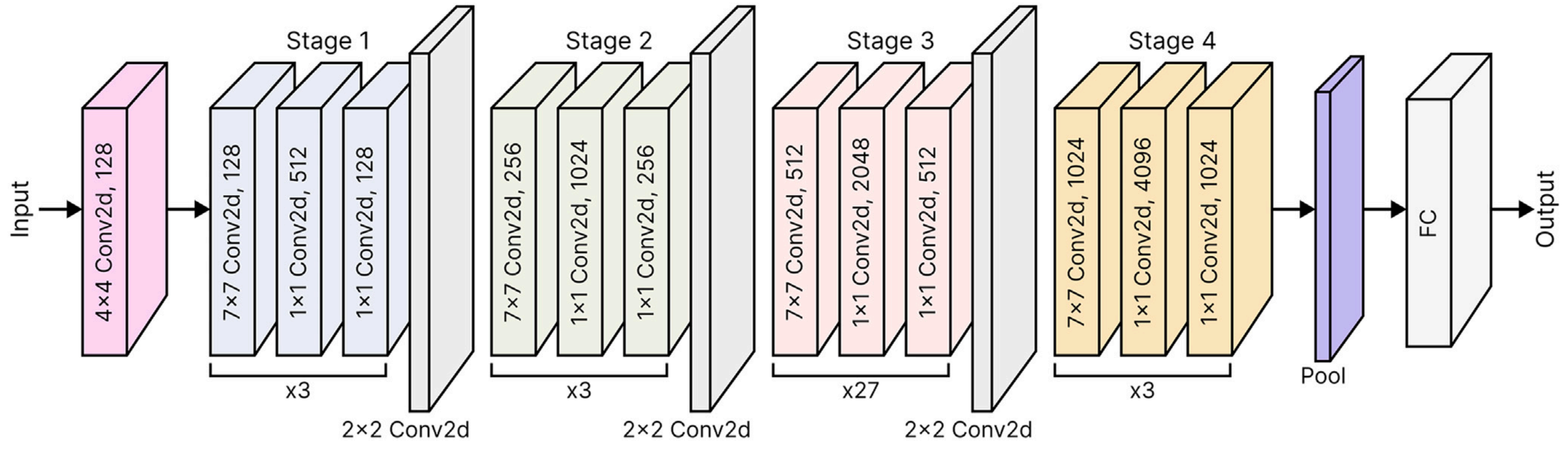

**Fig. 5.** ConvNeXt architecture structure (base version). Stages are connected sequentially, where the output of each stage serves as the input to the subsequent stage through downsampling operations.

**Table 1**
Details of ConvNeXt variants.

| Model | Resolution (pixels) | Channels (per stage) | Depths (per stage) | Params (M) | FLOPs (G) |
|---|---|---|---|---|---|
| ConvNeXt-tiny | 224x224 | [96, 192, 384, 768] | [3, 3, 9, 3] | 28 | 4.5 |
| ConvNeXt-small | 224x224 | [96, 192, 384, 768] | [3, 3, 27, 3] | 50 | 8.7 |
| ConvNeXt-base | 224x224 | [128, 256, 512, 1024] | [3, 3, 27, 3] | 89 | 15.4 |
| ConvNeXt-large | 224x224 | [192, 384, 768, 1536] | [3, 3, 27, 3] | 198 | 34.4 |

In the 4-channel configuration, the complementary nature of RGB and NIR bands is exploited, where RGB provides detailed spatial and chromatic information, while NIR enhances the detection of vegetation and other materials with strong near-infrared reflectance [57]. This configuration serves as a balanced and efficient baseline for multispectral feature extraction. In the 6-channel setting, the inclusion of NDVI and NDWI aims to provide a more explicit encoding of spectral relationships, facilitating the differentiation of subtle land-cover patterns. NDVI emphasizes vegetation by leveraging the contrast between red and near-infrared reflectance [18,20,25,74], while NDWI enhances the representation of water bodies through the contrast between near-infrared and green reflectance [21,50,51,60]. NDVI and NDWI indices are calculated from the available image bands, as defined in Eqs. (1) and (2), respectively.

$$\mathrm{NDVI} = \frac{\mathrm{NIR} - \mathrm{Red}}{\mathrm{NIR} + \mathrm{Red}} \tag{1}$$

$$\mathrm{NDWI} = \frac{\mathrm{Green} - \mathrm{NIR}}{\mathrm{Green} + \mathrm{NIR}} \tag{2}$$

The selection of NDVI and NDWI is not arbitrary, but rather grounded in extensive prior work, where these indices are among the most widely adopted spectral descriptors in land-cover and land-use classification. As reflected in numerous studies [22,42,50,54,70,73], NDVI and NDWI are consistently employed to enhance the discrimination of vegetation and water-related classes, respectively. By incorporating these indices, the models benefit from additional spectral information that complements the RGB and NIR channels, leading to a more robust and accurate feature representation for LCC.

In addition, it is important to note that both datasets employed in this study contain a relatively rich set of vegetation- and water-related semantic classes with high intra-category variability and inter-class similarity, especially for FBP. This characteristic further motivates the inclusion of NDVI and NDWI as representative spectral descriptors, as they are particularly effective at capturing subtle spectral variations within these complex categories.

While other spectral indices related to built-up areas, impervious surfaces, or additional land-cover types do exist, the choice of NDVI and NDWI in this work is guided by their simplicity, physical interpretability, the nature of the evaluated datasets, and their widespread adoption in land-cover classification studies. Their inclusion therefore reflects a well-grounded design choice, while the analysis of additional indices falls outside the scope of the present study and is acknowledged as a potential direction for future investigation.

### 3.2.2. Decoding and feature fusion

Once the visible and non-visible information has been processed through their respective encoder branches, the decoding process begins. Each branch independently reconstructs spatial details from the encoded feature maps, progressively recovering resolution through a series of decoder blocks. The decoding branches operate in parallel, maintaining the integrity of both visible and non-visible data streams. This separation ensures that the unique characteristics of each data source are preserved, while enabling a robust reconstruction of the input information at multiple scales.

The decoding process is based on a Feature Pyramid Network (FPN) [40], which systematically reconstructs spatial details from lower to higher resolutions. This pyramid-based approach enables the model to combine multi-scale feature representations, progressively refining the output as it transitions through each decoding stage. The decoding occurs in five stages, each focusing on recovering spatial resolution from the preceding lower-resolution feature maps. Starting with the most abstract and compact features from the encoder's output, each stage progressively refines and reconstructs these details. This process ensures that critical features are effectively preserved and enhanced, enabling the final reconstruction to align with the input resolution while maintaining strong discriminative capabilities.

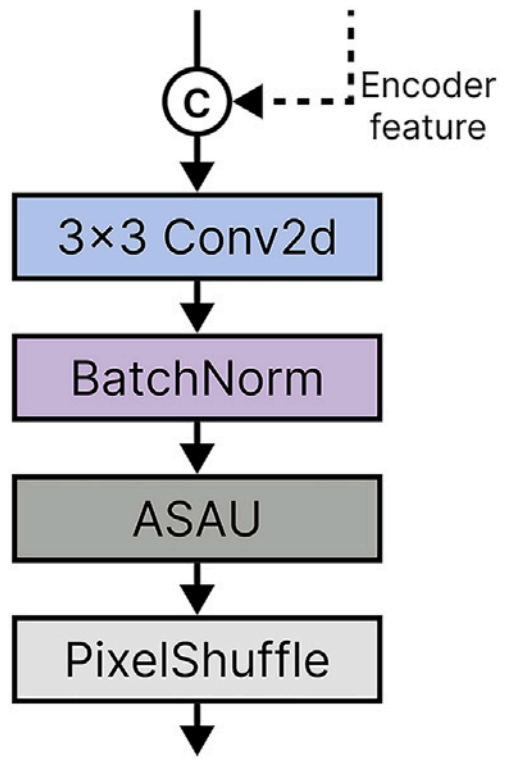

**Fig. 6.** Structure of the employed decoder block.

Each decoder block is composed of four components, as shown in Fig. 6. It is designed to progressively enhance and upscale the feature maps received from the encoder and previous decoding stages. Initially, a convolutional layer with a 3 × 3 kernel and appropriate padding adjusts the channel dimensions while preserving spatial resolution. This is followed by batch normalization, which stabilizes training by normalizing feature distributions. Next, a key design aspect in our design is the integration of the Adaptive Smooth Activation Unit (ASAU) [3], which is an activation function that provides a smooth approximation to the maximum operator. ASAU is a recently proposed function originally developed for classification and segmentation of complex medical structures. Its formulation combines softplus and learnable parameters to enable more stable and informative gradient flow during training, yielding continuous and differentiable transitions even at points where standard functions like ReLU and Leaky ReLU are non-differentiable.

Given the mentioned characteristics, ASAU is particularly suitable for segmentation tasks that require fine-grained spatial precision and sensitivity to subtle variations. We include it in our architecture to enhance the model's ability to capture fine textural differences and class boundaries, ultimately contributing to more accurate segmentation results. Compared to the original ASAU, which uses a fixed residual weight $w_0 = 0.01$ and learnable parameters initialized to $w_1 = 1.0$, and $w_2 = 1.0$, we introduce modified initializations $w_0 = 0.05$, $w_1 = 0.5$, and $w_2 = 1.5$, making $w_0$ also learnable. This particular combination of values was selected based on empirical testing, showing favorable convergence and segmentation behavior during early experiments. Increasing $w_0$ enhances the residual (identity) contribution during early training, mitigating vanishing gradients. Reducing $w_1$ lowers the initial gain of the nonlinear path, while increasing $w_2$ steepens its response, enabling sharper activation dynamics. These adjustments aim to improve convergence and allow the activation to adapt more flexibly to the characteristics of multispectral inputs. The comparison between the two versions of ASAU can be seen in Fig. 7.

Finally, a pixel shuffle [65] operation rearranges the feature map channels to upsample (i.e., double) the spatial resolution, enabling a smooth transition to higher-resolution stages. Additionally, each decoder block incorporates skip connections from the encoder, allowing access to both detailed low-level features and high-level abstract representations learned in deeper layers of the network.

Once the decoding branches begin reconstructing spatial details from their respective inputs, the architecture introduces a dedicated FPN-based fusion branch positioned between the two decoding streams (see Fig. 3). Instead of using simple concatenation to combine the outputs of each decoder branch, which can lead to suboptimal integration of

**Listing 1** Original ASAU

```
class ASAU(nn.Module):
 def __init__(self):
  super(ASAU,self).__init__()
  self.w1 = nn.Parameter(torch.tensor(1.0))
  self.w2 = nn.Parameter(torch.tensor(1.0))

 def forward(self,x):
  self.w0 = 0.01
  return self.w0 * x + ((1.0-self.w0) *
  x * torch.tanh(self.w2 * F.softplus(
  (1.0-self.w0) * self.w1 * x)))
```

**Listing 2** Modified ASAU

```
class ASAU(nn.Module):
 def __init__(self):
  super(ASAU,self).__init__()
  self.w0 = nn.Parameter(torch.tensor(0.05))
  self.w1 = nn.Parameter(torch.tensor(0.5))
  self.w2 = nn.Parameter(torch.tensor(1.5))

 def forward(self,x):
  return self.w0 * x + ((1.0-self.w0) *
  x * torch.tanh(self.w2 * F.softplus(
  (1.0-self.w0) * self.w1 * x)))
```

**Fig. 7.** Comparison of the original ASAU formulation and the modified ASAU version used in this work.

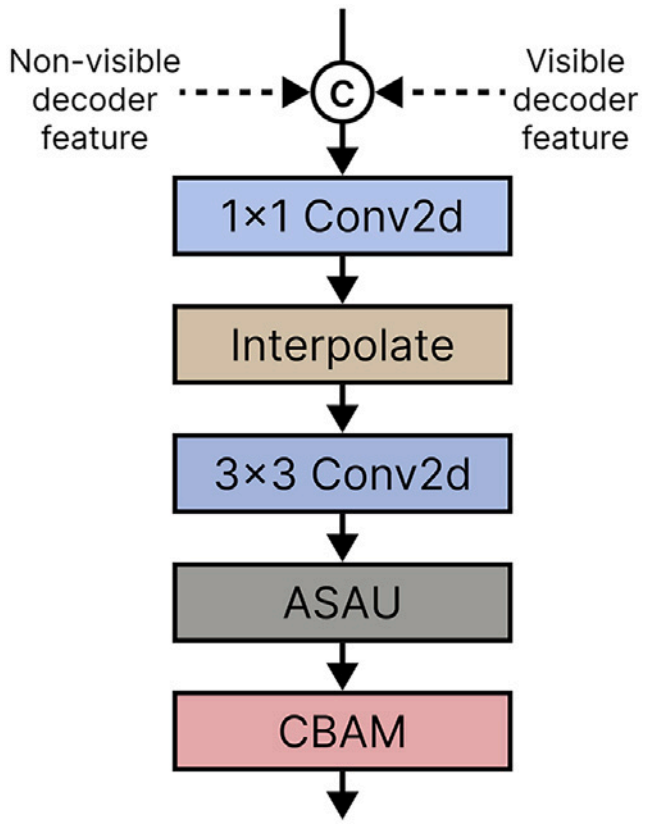

**Fig. 8.** Structure of the employed fusion block.

features due to the lack of explicit refinement [89], this branch progressively combines and refines features at each decoding stage. This approach effectively integrates low-level spatial details with high-level semantic information, ensuring a more coherent and discriminative reconstruction.

The fusion process is structured into four distinct stages, with each stage employing a dedicated fusion block, as shown in Fig. 8. At each fusion stage, the features from the corresponding decoding stages of both branches are first concatenated to form a unified representation. This concatenated output is then passed through a series of operations within the fusion block. The process begins with a $1 \times 1$ convolution, which aligns the feature maps in terms of dimensionality. Next, the features are upsampled using interpolation and combined through addition to reconstruct the spatial resolutions. Following this, a $3 \times 3$ convolution is used to refine the integrated features and improve spatial coherence. ASAU is then applied to increase the representational capacity of the fused features and better capture local variations introduced during the merging of both branches.

Before the fused features are passed to the next fusion stage, the architecture employs a Convolutional Block Attention Module (CBAM) [75] to recalibrate the features. This recalibration step is crucial for enhancing the quality of the fused representations, as it allows the model to focus on the most informative regions in both the spatial and channel dimensions [52]. The decision to include CBAM is based on its proven effectiveness in selectively emphasizing relevant features while suppressing redundant information [27], thereby improving the overall discriminative capacity of the model.

CBAM is a hybrid attention mechanism that integrates two complementary strategies: channel attention and spatial attention, as shown in Fig. 9. This mechanism first focuses on recalibrating the significance of individual channels to prioritize the most informative features. Subsequently, it applies spatial attention to highlight the most critical regions within the image, ensuring that both feature importance and spatial relevance are effectively captured [15,78]. By integrating CBAM, the architecture ensures that the fused features are better aligned with the underlying patterns of the data, facilitating a more meaningful and effective fusion process in subsequent stages.

After completing the four stages of fusion, the architecture employs an additional block dedicated to the final refinement of the feature maps. Structurally, this final fusion block consists of three components arranged sequentially: a $3 \times 3$ convolutional layer, which reduces the combined feature maps from all stages into a single, followed by batch normalization and a ReLU activation function. This block consolidates the outputs from the fusion stages into a unified representation, ultimately generating the segmentation output.

### 3.3. *Performance measurement*

To evaluate the performance of the proposed model, we adopt a set of metrics widely used in the semantic segmentation literature. These metrics provide complementary perspectives on model behavior and are particularly suited for multi-class scenarios. Each metric is described below.

#### 3.3.1. *Overall accuracy*

Overall Accuracy (OA) evaluates the fraction of correctly classified pixels relative to the total number of pixels in an image. Higher values ($\text{OA} \in [0,1]$) indicate that the model effectively segments the image, accurately differentiating between the various classes. The mathematical expression for OA is provided in Eq. (3):

$$\text{OA} = \frac{TP + TN}{TP + TN + FP + FN}, \tag{3}$$

where $TP$, $TN$, $FP$, and $FN$ represent true positives, true negatives, false positives, and false negatives, respectively.

#### 3.3.2. *Intersection over union*

Intersection over Union (IoU) is a critical metric for evaluating segmentation tasks. It quantifies the degree of overlap between the predicted segmentation mask and the ground-truth, serving as a direct measure of the model's ability to accurately delineate target regions. Models achieving higher IoU values indicate superior segmentation accuracy, with $\text{IoU} \in [0,1]$. The mathematical formulation for IoU is provided in Eq. (4):

$$\text{IoU} = \frac{|A \cap B|}{|A \cup B|}, \tag{4}$$

where $|A \cap B|$ denotes the pixel count correctly classified as part of the target class in both the prediction and ground-truth, while $|A \cup B|$ indicates the total number of pixels identified as part of the target class in either the prediction or the ground-truth.

#### 3.3.3. *Mean intersection over union*

Mean Intersection over Union (mIoU) extends the IoU metric by averaging the IoU scores calculated for each class. This provides a comprehensive measure of the model's performance across all categories in the dataset. A higher mIoU score, with mIoU $\in [0,1]$, reflects consistent segmentation accuracy across diverse classes. The computation of mIoU is detailed in Eq. (5):

$$\text{mIoU} = \frac{1}{C}\sum_{i=1}^{C} \text{IoU}_i, \tag{5}$$

where $\text{IoU}_i$ represents the IoU value for class $i$, and $C$ denotes the total number of classes in the dataset.

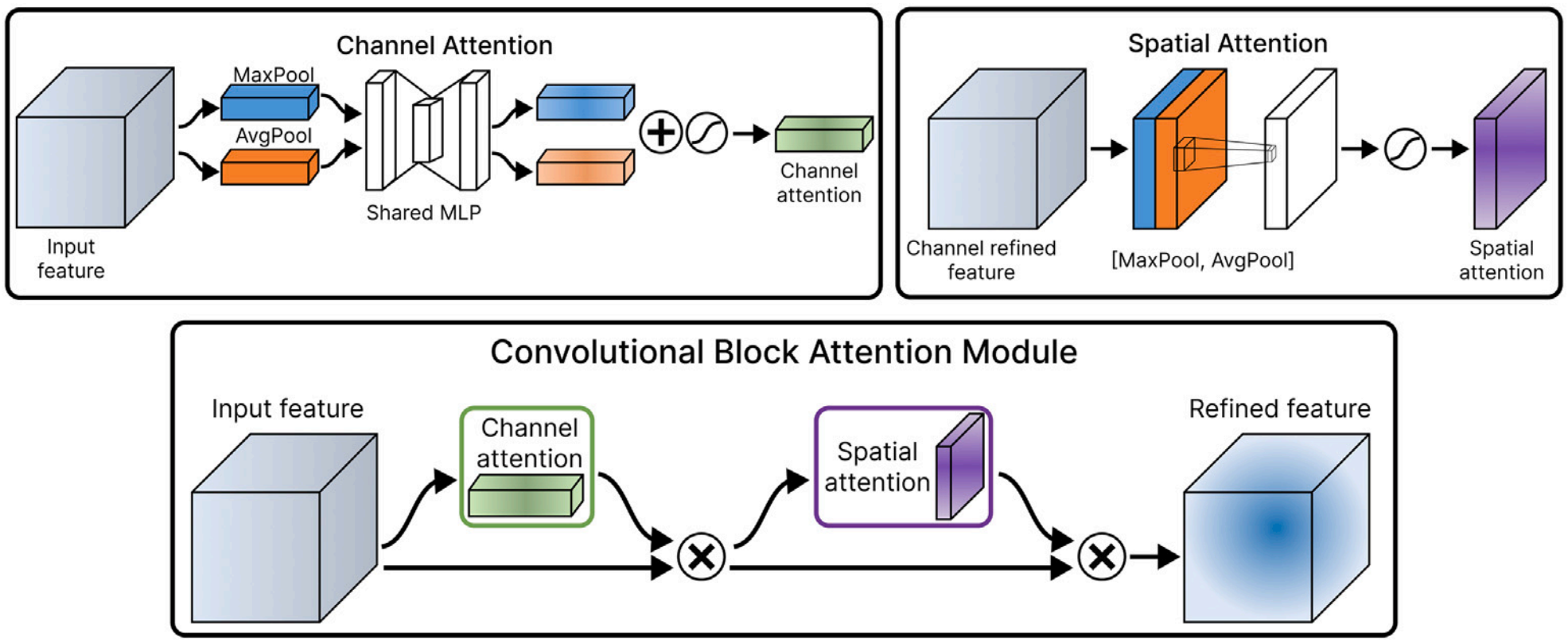

**Fig. 9.** CBAM mechanism and its components. ⊕ denotes element-wise summation, ⊗ denotes element-wise multiplication (attention gating), and AvgPool and MaxPool indicate average and max pooling operations, respectively.

### 3.3.4. Mean F1-score

The mean F1 (mF1) metric computes the average F1-score across all classes, offering a class-balanced evaluation in multi-class segmentation tasks [80]. It is particularly effective in scenarios with class imbalance, as it reflects both the precision and recall per class. Given $C$ classes, it is defined as shown in Eq. (6):

$$\text{mF1} = \frac{1}{C}\sum_{i=1}^{C}\left(2 \times \frac{\text{Precision}_i \times \text{Recall}_i}{\text{Precision}_i + \text{Recall}_i}\right), \tag{6}$$

with mF1 $\in [0,1]$. Higher values indicate better class-balanced performance.

## 3.4. Training and implementation details

In this work, the Potsdam dataset was divided into 28 tiles assigned to training and 10 tiles assigned to evaluation, as detailed in Table 3. For the FBP dataset, the split consisted of 120 tiles for training and 30 for evaluation, adhering to the division provided by the original authors [69]. Prior to training, the images were cropped into non-overlapping patches of $256 \times 256$ pixels. Each dataset was normalized using Min-Max scaling based on the statistics of the corresponding dataset, ensuring that all channels, including NDVI and NDWI, were mapped consistently. We also applied horizontal and vertical flips with a probability of 0.5, as well as random 90-degree rotations with the same probability for data augmentation.

The architecture was implemented in Python using the PyTorch framework. The ConvNeXt backbone was sourced from the TorchVision model library,[2] using pre-trained weights. For benchmarking, we also implemented widely used segmentation models such as U-Net, DeepLabV3+, and SegFormer. Baselines were implemented using their standard configurations, only adjusting their input layers to accommodate multispectral data when necessary. Testing was performed using 5-fold cross-validation to ensure robust evaluation. All experiments were conducted on a workstation equipped with two Nvidia A100 SXM4 GPUs with 40 GB of memory, supported by 64 CPU cores and 128 GB of RAM.

**Table 3**
Division of the training and evaluation subsets of the Potsdam dataset.

| Subset | Image ID |
|---|---|
| Training | 2_10, 2_11, 2_12, 3_10, 3_11, 3_12, 4_10, 4_11, 4_12, 4_15, 5_10, 5_11, 5_12, 5_15, 6_7, 6_8, 6_9, 6_10, 6_11, 6_12, 6_14, 6_15 7_7, 7_8, 7_9, 7_10, 7_11, 7_12 |
| Evaluation | 2_13, 2_14, 3_13, 3_14, 4_13, 4_14, 5_13, 5_14, 6_13, 7_13 |

Training was conducted under a consistent configuration for both the proposed and baseline models. We used a combined loss function composed of standard cross-entropy and Dice loss, weighted as: CE + $0.5 \cdot$ Dice. Optimization was performed using the AdamW optimizer with a learning rate of $1e-4$, weight decay of $1e-5$, and a batch size of 128. Models were trained for 150 epochs using mixed precision to reduce memory usage and accelerate training. We employed the OneCycleLR learning rate scheduler with cosine annealing, configured with $3e-4$ as the maximum learning rate, a warm-up phase covering 5% of the total epochs, a division factor of 10.0, and a final division factor of $1e3$. These hyperparameters were obtained via hyperparameter tuning using the HyperBand method from the Ray library.[3]

# 4. Results and discussion

## 4.1. Results on Five-Billion-Pixels dataset

Table 4 presents the performance of our proposed architecture across different encoder variants evaluated on two input configurations: RGB-NIR (4 channels) and RGB-NIR-NDVI-NDWI (6 channels). We observe a general performance progression as the encoder complexity increases from tiny to base. However, this trend does not hold for the large variant, which, despite being the most complex, slightly underperforms compared to the base version. We argue that this outcome may stem from the large model's substantially higher parameter count. Specifically, MeCSAFNet-large employs two ConvNeXt-large encoders, each with 198M parameters, meaning that a single encoder in the large variant has more than double the parameters of a ConvNeXt-base encoder. Evidently, such a configuration requires more epochs to reach full convergence.

Due to the above factors, and given the considerable dataset volume (150 large tiles in the FBP dataset), the fixed training schedule

[2] https://pytorch.org/vision/stable/models.html

[3] https://docs.ray.io/en/latest/tune/api/doc/ray.tune.schedulers.HyperBandScheduler.html

**Table 4**
Performance of the different MeCSAFNet variants on the Five-Billion-Pixels dataset. The best results for each spectral group are underlined.

| Bands | Model | OA (%) | mIoU (%) | mF1 (%) | T. time (h) | Speed (s) |
|---|---|---|---|---|---|---|
| RGB NIR | MeCSAFNet-tiny | 90.21 | 69.18 | 79.92 | 10.05 | 0.0319 |
| | MeCSAFNet-small | 90.47 | 71.08 | 81.13 | 14.46 | 0.0513 |
| | MeCSAFNet-base | 90.72 | 71.23 | 81.45 | 18.30 | 0.0509 |
| | MeCSAFNet-large | 90.54 | 70.53 | 81.04 | 19.86 | 0.0503 |
| RGB NIR NDVI NDWI | MeCSAFNet-tiny | 90.42 | 70.60 | 81.04 | 12.72 | 0.0324 |
| | MeCSAFNet-small | 90.59 | 71.40 | 81.86 | 16.90 | 0.0512 |
| | MeCSAFNet-base | 91.79 | 72.79 | 82.54 | 19.80 | 0.0507 |
| | MeCSAFNet-large | 90.52 | 71.69 | 81.82 | 22.36 | 0.0508 |

**Table 5**
Performance comparison of our approach with other baseline methods in the evaluation set of the Five-Billion-Pixels dataset (30 tiles/21,840 patches). Note that we have selected the best MeCSAFNet models from each spectral configuration for comparison. Speed refers to the average inference time per image patch. The best results for each spectral group are underlined.

| Model | IoU (%) | | | | | | | | | | | | | | | |
|---|---|---|---|---|---|---|---|---|---|---|---|---|---|---|---|---|
| | Indu | Padd | Irri | Dryc | Gard | Arbo | Shru | Park | Natu | Arti | River | Urba | Lake | Pond | Fish | Snow |
| U-Net (4c) | 76.68 | 77.67 | 88.15 | 33.52 | 34.16 | 94.86 | 28.68 | 39.32 | 89.31 | 52.25 | 64.04 | 82.20 | 73.75 | 17.99 | 79.64 | 51.16 |
| SegFormer (4c) | 74.08 | 77.12 | 88.09 | 33.70 | 35.65 | 95.19 | 27.05 | 56.79 | 90.32 | 46.70 | 66.41 | 80.10 | 77.88 | 20.79 | 78.49 | 40.68 |
| DeepLabV3+ (4c) | 76.69 | 78.56 | 87.84 | 31.14 | 36.71 | 96.30 | 36.26 | 53.51 | 90.41 | 59.81 | 64.26 | 81.13 | 76.38 | 29.88 | 78.94 | 68.04 |
| **MeCSAFNet-base** (4c) | 80.12 | 83.42 | 90.07 | 44.74 | 45.52 | 97.50 | 43.63 | 56.06 | 94.01 | 67.31 | 79.19 | 83.65 | 83.20 | 24.98 | 88.28 | 86.48 |
| U-Net (6c) | 77.00 | 77.33 | 90.18 | 52.53 | 31.65 | 95.53 | 32.80 | 51.09 | 90.04 | 47.00 | 68.55 | 81.76 | 78.84 | 23.22 | 81.25 | 74.98 |
| DeepLabV3+ (6c) | 74.11 | 81.72 | 92.77 | 67.32 | 44.08 | 96.20 | 50.58 | 48.76 | 91.00 | 65.19 | 72.27 | 80.01 | 79.89 | 27.25 | 84.76 | 56.44 |
| SegFormer (6c) | 74.22 | 73.62 | 89.26 | 43.61 | 30.36 | 95.76 | 33.39 | 60.77 | 89.02 | 59.88 | 64.85 | 80.12 | 75.83 | 17.82 | 77.84 | 64.16 |
| **MeCSAFNet-base** (6c) | 79.06 | 83.00 | 94.19 | 74.47 | 48.93 | 97.14 | 40.23 | 64.66 | 93.74 | 67.59 | 76.52 | 83.44 | 82.09 | 21.84 | 86.69 | 90.54 |

| Model | IoU (%) | | | | | | | | mIoU (%) | mF1 (%) | OA (%) | T. time (h) | Speed (s) |
|---|---|---|---|---|---|---|---|---|---|---|---|---|---|
| | Bare | Rura | Stad | Squa | Road | Over | Rail | Airp | | | | | |
| U-Net (4c) | 87.27 | 79.11 | 50.06 | 22.60 | 77.76 | 69.84 | 36.51 | 58.81 | 61.06 | 73.04 | 87.34 | 9.37 | 0.0207 |
| SegFormer (4c) | 85.96 | 77.63 | 52.54 | 18.08 | 76.28 | 68.92 | 37.75 | 54.20 | 60.85 | 72.88 | 87.70 | 7.81 | 0.0175 |
| DeepLabV3+ (4c) | 88.07 | 77.68 | 49.42 | 26.73 | 75.73 | 67.89 | 39.63 | 49.79 | 63.37 | 75.41 | 87.65 | 8.23 | 0.0194 |
| **MeCSAFNet-base** (4c) | 93.43 | 81.05 | 59.02 | 38.58 | 78.46 | 76.28 | 55.49 | 79.05 | 71.23 | 81.45 | 90.72 | 18.30 | 0.0509 |
| U-Net (6c) | 86.87 | 78.20 | 50.67 | 12.01 | 76.44 | 68.47 | 37.11 | 59.32 | 63.45 | 74.92 | 89.03 | 13.70 | 0.0213 |
| DeepLabV3+ (6c) | 81.44 | 78.55 | 45.99 | 3.16 | 73.93 | 71.70 | 41.16 | 80.11 | 66.18 | 77.04 | 89.19 | 12.54 | 0.0250 |
| SegFormer (6c) | 86.79 | 77.40 | 52.46 | 25.83 | 75.71 | 68.45 | 43.05 | 62.40 | 63.44 | 75.36 | 87.89 | 14.26 | 0.0182 |
| **MeCSAFNet-base** (6c) | 94.32 | 81.45 | 57.13 | 37.98 | 78.50 | 76.62 | 55.51 | 81.28 | 72.79 | 82.54 | 91.79 | 22.36 | 0.0508 |

The abbreviations of the categories are as follows: Indu - industrial area, Padd - paddy field, Irri - irrigated field, Dryc - dry cropland, Gard - garden land, Arbo - arbor forest, Shru - shrub forest, Natu - natural meadow, Arti - artificial meadow, Urba - urban residential, Fish - fish pond, Bare - bare land, Rura - rural residential, Stad - stadium, Squa - square, Over - overpass, Rail - railway station, Airp - airport.

may have constrained the model's learning, preventing the large variant from reaching its full potential. Nonetheless, the large variant still surpasses the tiny and small models in certain metrics. For instance, the 4-channel model achieves higher OA, and the 6-channel version shows better mIoU. This suggests that extended training could enable it to outperform the base model.

Notably, MeCSAFNet-base stands out as the most balanced and effective variant across both input settings, achieving the best overall results. In the 4-channel configuration, this variant achieves an OA of 90.72% and a remarkable mIoU of 71.23%, demonstrating better surface and edge delineation capabilities than the other variants. Likewise, the 6-channel configuration records the highest OA (91.79%), mF1 (82.54%), and mIoU (72.79%) among all. Broadly, the inclusion of spectral indices as additional input channels is associated with a consistent improvement in segmentation performance across all variants.

In terms of training time, all variants require several hours, which is expected considering the dataset's scale. The 6-channel versions tend to take longer, reflecting the added complexity from the additional input features. However, the increase remains reasonable and proportional to the model size and input configuration. Moreover, the inference speed remains fast across all variants, with no configuration exceeding 0.06 s per patch. This reinforces the practicality of our architecture, showing that it maintains low-latency inference even when equipped with heavier encoders and richer spectral inputs. This underscores the efficiency of MeCSAFNet's design, supporting its applicability in real-world scenarios.

Table 5 presents a comparison between our model and different segmentation baselines using the FBP dataset. Our method, in both 4-channel and 6-channel configurations, clearly outperforms U-Net, DeepLabV3+, and SegFormer across all evaluation metrics. None of the baseline models surpass 70% in mIoU, 80% in mF1, or 90% in OA, even when extended to 6-channel input. In contrast, MeCSAFNet-base exceeds all these thresholds in both channel settings, confirming its superior ability to accurately classify pixels, delineate boundaries, and distinguish complex land cover types. Performance gaps are especially pronounced in challenging classes such as snow, where the 4-channel and 6-channel versions of our model reach 86.48% and 90.54% IoU, respectively. These results are significantly higher than those of all baseline models, none of which exceed 70% in this class. This highlights the model's robustness in scenarios where other architectures struggle.

In terms of efficiency, our models are the slowest to train, both reaching around 20 hours. However, it is important to clarify that this comparison includes the best-performing MeCSAFNet variants, and in both cases, they correspond to the base version, which naturally requires more time due to its parameter count. Notably, even the lighter MeCSAFNet-tiny (4c), outperforms all baselines across all metrics (see Table 4), while maintaining a training time comparable to them. This confirms that the advantage of our approach is not limited to large models but holds even in configurations with similar computational cost. Regarding inference, our models also present the highest latency, which is expected given the multi-branch structure. Nevertheless, inference time remains low, with a maximum of 0.0509 s per patch, making the models viable for practical deployment.

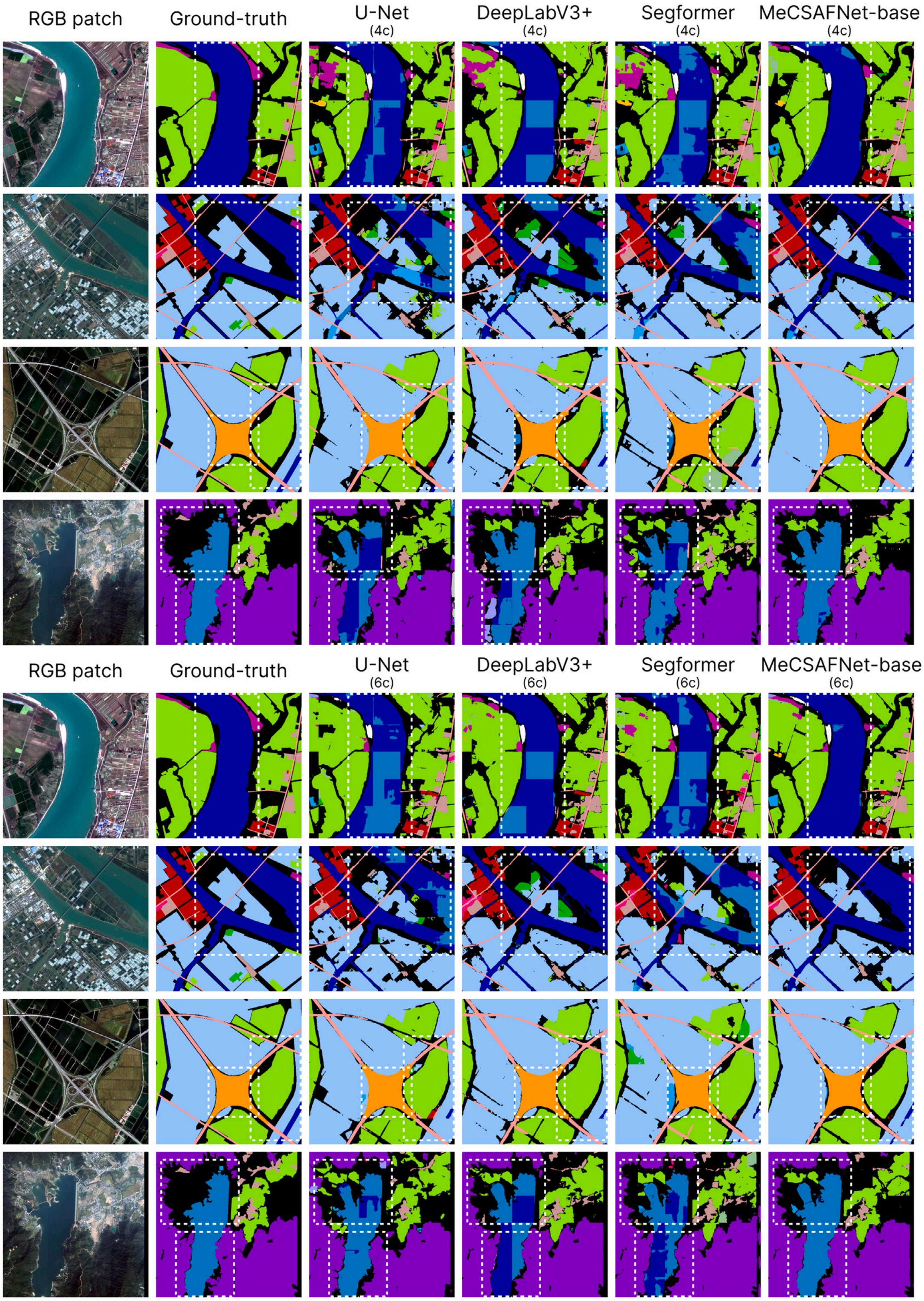

**Fig. 10.** Segmentation results of our approach compared to other baseline methods on the Five-Billion-Pixels dataset. Regions demarcated by dashed lines indicate the areas where the most substantial discrepancies between models are observed.

For a more in-depth analysis, Fig. 10 presents a visual comparison of the segmentation results across baseline models and our approach, using both 4-channel and 6-channel configurations. The qualitative results align with the numerical findings, showing that our model produces more complete region labeling and fewer misclassifications. It consistently separates adjacent land cover types more effectively, maintaining clearer boundaries even in complex scenes. A recurring issue in the baselines is the confusion between similar classes such as river, lake, and pond, often generating fragmented or mixed regions. In contrast, our model accurately isolates these classes, preserving the semantic structure of the scene. Moreover, the 6-channel version of MeCSAFNet further improves the segmentation by refining class

**Table 6**
Comparison of our approach with other state-of-the-art methods on the Five-Billion-Pixels dataset. The best results are underlined.

| Method | OA (%) | mIoU (%) | mF1 (%) | Year | Ref. |
|---|---|---|---|---|---|
| Advent | 74.81 | 38.98 | – | 2019 | [72] |
| SimCLR | 64.31 | 21.34 | – | 2020 | [7] |
| FALSE | 64.88 | 21.41 | – | 2022 | [86] |
| DPA + U-Net | 80.35 | 44.51 | 57.34 | 2023 | [69] |
| Class-Aware ResUNet | 75.97 | 40.68 | – | 2023 | [9] |
| DPA + DeepLabv3 + | 79.87 | 42.12 | 54.84 | 2023 | [69] |
| GFCNet | 65.44 | 21.84 | – | 2023 | [85] |
| GeoAgent + DeepLabv3 + | – | 62.25 | 63.21 | 2023 | [45] |
| Mix Transformer | 78.30 | 52.78 | 60.96 | 2024 | [5] |
| U-Net + SK-ResNeXt50 | 80.56 | 54.39 | – | 2025 | [59] |
| ConvNeXt + Transformer | 88.10 | – | – | 2025 | [35] |
| CSNet | 90.50 | 70.30 | – | 2025 | [29] |
| **MeCSAFNet-base (4c)** | 90.72 | 71.23 | 81.45 | 2026 | ours |
| **MeCSAFNet-base (6c)** | 91.79 | 72.79 | 82.54 | 2026 | ours |

*Dash (–) indicates metrics not reported in the original publication.*

boundaries and filling in previously incomplete areas, particularly in cluttered environments. In contrast, some baselines such as SegFormer show performance degradation when using the extended spectral input, leading to noisier predictions and weaker class separation.

Finally, Table 6 compares our approach with recent state-of-the-art methods on the FBP dataset. MeCSAFNet, in both 4-channel and 6-channel configurations, outperforms all prior works in OA, mIoU, and mF1-score. Specifically, MeCSAFNet-base (6c) achieves the best performance among all, reaching 91.79% OA, 72.79% mIoU, and 82.54% mF1. This model surpasses others based on traditional architectures, such as U-Net and DeepLabV3 + , as well as those integrating enhancement modules like Selective Kernel (SK)-ResNeXt or GeoAgent. Our approach also surpasses transformer-based methods, including Mix Transformer. Notably, CSNet and U-Net + SK-ResNeXt achieve strong performance, yet fall short in mIoU and OA compared to our results. These results confirm that MeCSAFNet delivers consistent gains across both global and class-wise segmentation metrics, without relying on additional supervision or complex components.

### *4.2. Results on potsdam dataset*

To begin with, Table 7 reports the performance of different MeCSAFNet variants on the Potsdam dataset. In the 4-channel configuration, a consistent improvement in performance is observed as model complexity increases, with the large variant achieving the best results (91.17% OA, 84.16% mIoU, and 91.27% mF1). This behavior can be attributed to the relatively lower task complexity and reduced training sample diversity, which allow higher-capacity models to effectively converge within the fixed training budget. In contrast, for the 6-channel configuration, the best-performing model is MeCSAFNet-base, with 91.18% OA, 84.14% mIoU, and 91.24% mF1, while the large variant does not surpass it. We attribute this behavior to the combined effect of increased architectural complexity and higher spectral dimensionality, which leads to a more challenging optimization problem that may require additional training epochs for full convergence. Nevertheless, the large variant still outperforms the tiny and small models across all metrics, indicating a clear upward performance trend with increasing model capacity.

Regarding efficiency, there is a clear trend of increasing training time as model complexity grows. In addition, all variants trained with the 6-channel configuration require more time than their 4-channel counterparts, as expected. Notably, the tiny variants in both configurations achieve strong results, with OA values above 90% and mIoU above 83%, while maintaining relatively short training times. Inference speed also remains low, particularly for the lighter models, while the more complex variants reach values just above 0.05 s per patch. These inference times are acceptable for use in practical scenarios.

Next, Table 8 compares our approach, selecting the best MeCSAFNet variant for each spectral configuration, against different baseline models on the Potsdam dataset. As shown in the OA column of Table 8, both MeCSAFNet variants exceed 91%, while none of the baseline models, including U-Net, DeepLabV3 + , and SegFormer, reach 89%. A similar pattern is observed in terms of mIoU, where MeCSAFNet-large (4c) and MeCSAFNet-base (6c) achieve 84.16% and 84.14%, respectively, whereas only SegFormer (6c) crosses the 80% mark among the baselines. These results indicate that our models are more effective in accurately identifying and segmenting image regions in this dataset. A per-class analysis of IoU further reinforces this conclusion. In the building class, which is particularly challenging due to high geometric variability and intra-class diversity, both MeCSAFNet variants surpass 93%, while no baseline reaches 90%. Similar advantages are seen in low vegetation and impervious surfaces, where our models consistently achieve higher IoU values than all competitors.

**Table 7**
Performance of the different MeCSAFNet variants on the Potsdam dataset. The best results for each spectral group are underlined.

| Bands | Model | OA (%) | mIoU (%) | mF1 (%) | T. time (h) | Speed (s) |
|---|---|---|---|---|---|---|
| RGB NIR | MeCSAFNet-tiny | 90.87 | 83.56 | 90.90 | 2.16 | 0.0310 |
| | MeCSAFNet-small | 90.91 | 83.63 | 90.97 | 2.90 | 0.0499 |
| | MeCSAFNet-base | 91.06 | 84.03 | 91.19 | 3.72 | 0.0497 |
| | MeCSAFNet-large | 91.17 | 84.16 | 91.27 | 5.26 | 0.0499 |
| RGB NIR NDVI NDWI | MeCSAFNet-tiny | 90.93 | 83.72 | 91.00 | 2.83 | 0.0494 |
| | MeCSAFNet-small | 90.97 | 83.84 | 91.08 | 3.27 | 0.0506 |
| | MeCSAFNet-base | 91.18 | 84.14 | 91.24 | 4.16 | 0.0500 |
| | MeCSAFNet-large | 91.13 | 84.07 | 91.21 | 5.91 | 0.0537 |

**Table 8**
Performance comparison of our approach with other baseline methods in the evaluation set of the Potsdam dataset (10 tiles/5,290 patches). Note that we have selected the best MeCSAFNet models from each spectral configuration for comparison. Speed refers to the average inference time per image patch. The best results for each spectral group are underlined.

| Model | IoU (%) | | | | | mIoU (%) | mF1 (%) | OA (%) | T. time (h) | Speed (s) |
|---|---|---|---|---|---|---|---|---|---|---|
| | Building | Low veg. | Tree | Car | Impervious | | | | | |
| U-Net (4c) | 87.25 | 70.79 | 71.77 | 83.68 | 80.82 | 78.86 | 88.03 | 87.69 | 1.53 | 0.0200 |
| DeepLabV3+ (4c) | 88.21 | 70.70 | 71.21 | 83.66 | 81.43 | 79.04 | 88.12 | 87.71 | 1.41 | 0.0194 |
| SegFormer (4c) | 84.42 | 71.39 | 70.32 | 79.84 | 79.66 | 77.13 | 86.98 | 88.88 | 1.35 | 0.0173 |
| **MeCSAFNet-large (4c)** | 93.44 | 77.23 | 76.76 | 86.70 | 86.66 | 84.16 | 91.27 | 91.17 | 5.26 | 0.0499 |
| U-Net (6c) | 84.86 | 71.37 | 73.09 | 84.58 | 82.37 | 79.25 | 88.68 | 88.26 | 1.72 | 0.0210 |
| DeepLabV3+ (6c) | 85.24 | 73.33 | 73.36 | 83.57 | 82.02 | 79.51 | 88.49 | 88.19 | 1.52 | 0.0201 |
| SegFormer (6c) | 87.61 | 73.36 | 73.82 | 83.75 | 82.97 | 80.30 | 88.96 | 88.76 | 1.49 | 0.0180 |
| **MeCSAFNet-base (6c)** | 94.13 | 77.47 | 75.77 | 86.52 | 86.79 | 84.14 | 91.24 | 91.18 | 4.16 | 0.0500 |

To complement the quantitative results, Fig. 11 shows the qualitative segmentation results of our approach compared to baseline models on the Potsdam dataset. These visual tests are consistent with the performance metrics, as our models demonstrate a clearer ability to distinguish and separate regions accurately. In particular, our approach produces fewer misclassified areas and reduces visual artifacts, especially in classes with small or compact instances. Moreover, it achieves better region completeness, avoiding gaps or partial coverage in classes such as building and clutter. The 6-channel configuration further enhances this effect, as MeCSAFNet-base exhibits more coherent and continuous segmentation compared to its 4-channel counterpart. In contrast, models such as DeepLabV3+ and SegFormer show a decline in performance under the 6-channel input, often producing fragmented predictions and failing to preserve the spatial integrity of certain regions.

Moving forward, Table 9 presents a comparison of our approach with recent state-of-the-art methods on the Potsdam dataset. Both MeCSAFNet-large (4c) and MeCSAFNet-base (6c) outperform the rest of the models across all reported metrics. Our methods achieve over 91% OA and over 84% mIoU, while most other approaches remain below these thresholds. The results highlight the generalizability and robustness of our design, which outperforms a wide range of architectural strategies. This includes U-Net-based models, methods incorporating auxiliary supervision such as edge information, and approaches inspired by transformer architectures. Notably, MeCSAFNet surpasses even more complex pipelines that combine multiple components, such as hybrid frameworks integrating U-Net, Transformer modules, and the Segment Anything Model (SAM).[4] While some approaches report strong performance in isolated metrics, MeCSAFNet is among the few that maintain high values across OA, mIoU, and mF1 simultaneously, indicating consistent performance at both global and class-wise levels.

Finally, Table 10 presents an ablation study of our proposed approach using the MeCSAFNet-tiny variant on the Potsdam dataset. We evaluate the impact of varying the attention mechanism (CBAM, including its Channel Attention Module (CAM) and Spatial Attention Module (SAM)) and the activation function (ReLU, ASAU) across both spectral configurations. Results show that the combination of CBAM and ASAU consistently yields the best performance, achieving OA and mF1 values above 90% in both configurations. In the 4-channel setting, CAM and SAM paired with ASAU achieve competitive results, indicating that simpler attention mechanisms remain viable when coupled with a strong activation function. In contrast, in the 6-channel case, CAM and SAM perform better when paired with ReLU, pointing to a possible interaction between activation stability and spectral redundancy under limited model capacity. Training times for all configurations remain within a narrow range of approximately 2 to 3 hours, and inference speed differences are mostly negligible. Overall, the results validate the design choices adopted in our final architecture and confirm that the selected components contribute meaningfully to performance improvements.

### *4.3. Design analysis and discussion*

As shown throughout the experimental evaluation, our approach consistently outperforms a wide range of competing methods, each based on distinct architectural paradigms. This consistent advantage can be attributed to the design of MeCSAFNet and its core components, which are intended to maximize segmentation accuracy while preserving computational efficiency. The network was conceived not only to extract rich spectral-spatial features, but also to remain lightweight enough for real-time deployment in practical applications.

Therefore, to analyze the performance gains of our approach over other methods, Table 11 reports the relative improvements achieved by our best models in each dataset when compared to a selection of baseline and state-of-the-art models. Compared to standard baselines such as U-Net, DeepLabV3+, and SegFormer, our models demonstrate substantial gains in all metrics, particularly in mIoU. For example, on the FBP dataset, MeCSAFNet-base (6c) improves mIoU by up to +19.62% over SegFormer (4c), and by +14.74% over SegFormer (6c). Similarly, on the Potsdam dataset, MeCSAFNet-large (4c) achieves a +4.80% improvement in mIoU over SegFormer (6c), +5.85% over DeepLabV3+ (6c), +6.19% over U-Net (6c), and +6.72% over U-Net (4c). These improvements reflect the ability of our architecture to better exploit the spectral richness of the input, and to preserve region-level consistency across diverse classes. While other methods also receive extended spectral inputs, their performance gains remain limited. In contrast, our dual-branch encoder explicitly separates the processing of spectral components, and the observed results suggest that this design is better suited to extract and integrate information from heterogeneous bands.

In addition to the encoder structure, we attribute part of these gains to the integration of CBAM in the decoding stage and the use of ASAU as an activation function. CBAM improves spatial and channel-level focus during upsampling, while ASAU ensures smoother and more informative gradient propagation compared to ReLU, particularly in scenes with complex boundaries or high intra-class variation. This combination enables even compact variants like MeCSAFNet-tiny to achieve strong performance, as seen in earlier experiments, and contributes to the robustness of our larger models under different spectral configurations.

When comparing against state-of-the-art methods specifically designed for multispectral or high-resolution remote sensing, the improvements remain consistent. For instance, on the FPB dataset, our model surpasses the hybrid architecture U-Net + SK-ResNeXt50 by +13.94% in OA and +33.82% in mIoU. This highlights that while multi-path convolutional modules such as SK improve representation capacity, our dual-branch structure with ConvNeXt and the enhanced decoding path with CBAM achieves better spectral decoupling without increasing inference cost. Similarly, the improvement over CSNet (+3.54% mIoU) suggests that our strategy is more effective at extracting and integrating spectral-spatial features than domain adaptation or mixup-based augmentation approaches.

[4] Please note that we refer to the Segment Anything Model from Meta AI, and not to the Spatial Attention Module used within CBAM.

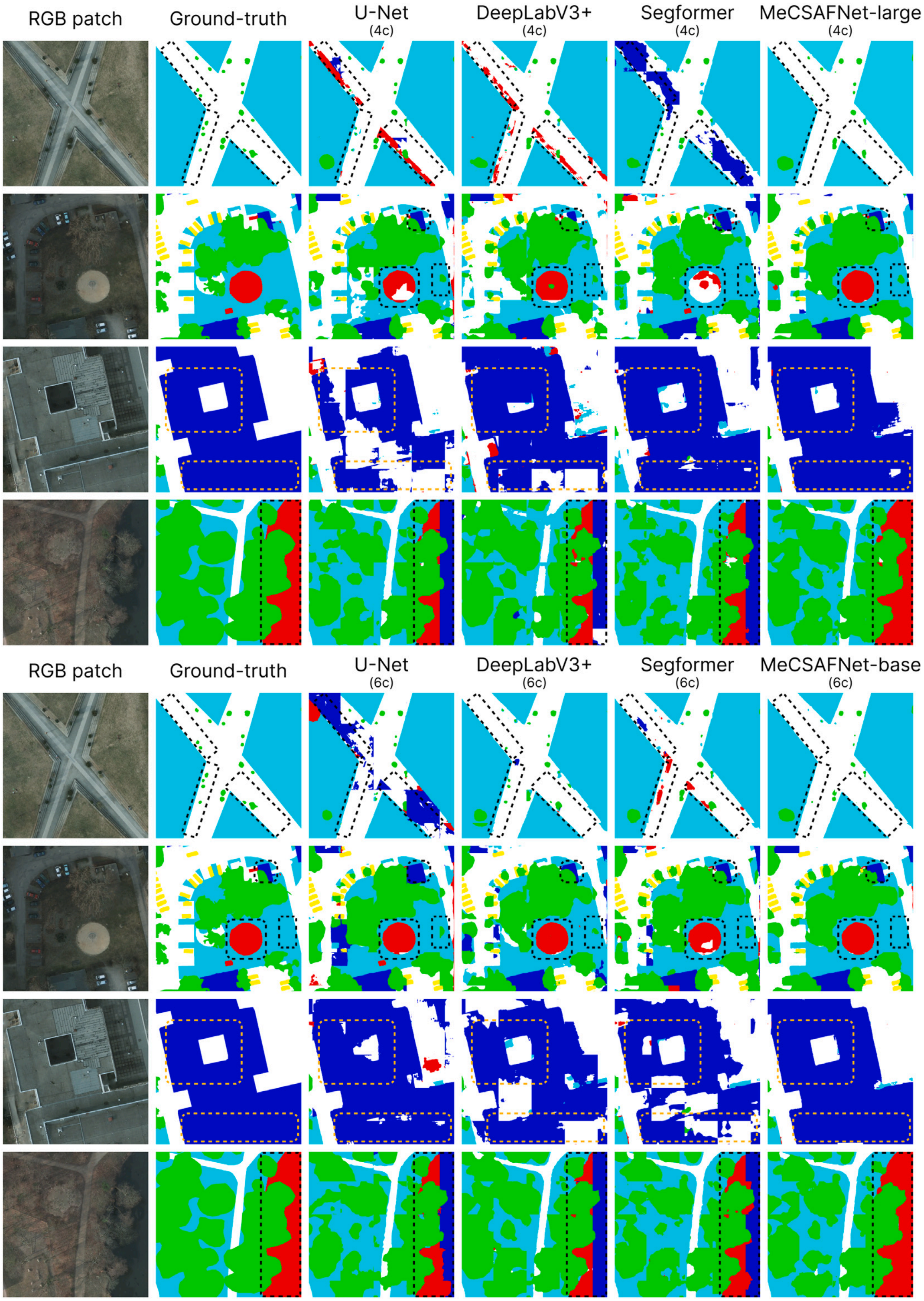

**Fig. 11.** Segmentation results of our approach compared to other baseline methods on the Potsdam dataset. Regions demarcated by dashed lines indicate the areas where the most substantial discrepancies between models are observed.

On the Potsdam dataset, MeCSAFNet-large (4c) also outperforms several recent state-of-the-art models. DSPCANet, one of the top-performing models, leverages digital surface model (DSM) data and attention-based fusion, while GFDNet integrates frequency-domain representations and complex hybrid attention. Despite these specialized mechanisms, our architecture achieves better segmentation accuracy, improving mIoU by +8.37% and +3.80% over DSPCANet and GFDNet, respectively. We attribute this to the ability of ConvNeXt to capture both fine-grained local patterns and broader spatial context through its large-kernel convolutions and hierarchical multi-stage design, together with the regularization introduced by the ASAU activation function, both integrated into

**Table 9**
Comparison of our approach with other state-of-the-art approaches on the Potsdam dataset. The best results are underlined.

| Method | OA (%) | mIoU (%) | mF1 (%) | Year | Ref. |
|---|---|---|---|---|---|
| ResNet-ASPP | 88.20 | 70.32 | – | 2021 | [6] |
| DSPCANet | 90.13 | 77.66 | 87.19 | 2021 | [37] |
| BAM-UNet | 89.13 | – | 88.59 | 2021 | [53] |
| BLASeNet | 79.64 | 62.90 | 75.47 | 2022 | [38] |
| MAResU-Net | 78.31 | 61.10 | – | 2022 | [34] |
| MSCR-HRNetV2 | 85.30 | 75.96 | – | 2023 | [30] |
| RegDA | – | 63.12 | – | 2024 | [43] |
| CxtHGNet | 87.15 | 70.28 | – | 2024 | [33] |
| CMPF-U-Net | 89.17 | 80.23 | – | 2024 | [32] |
| DPFE-AFF | 89.70 | 82.80 | 90.50 | 2024 | [28] |
| SCSNet | 89.43 | – | 90.51 | 2025 | [44] |
| AMView | 76.18 | 66.24 | 77.79 | 2025 | [64] |
| U-NetFormer + SAM | – | 75.50 | 84.81 | 2025 | [55] |
| FML-Swin | 85.63 | 78.58 | – | 2025 | [76] |
| Swin Transformer + Edge | – | 72.39 | 83.68 | 2025 | [41] |
| FGNet | 86.10 | 71.46 | 82.78 | 2025 | [84] |
| FILNet | 85.35 | 76.12 | – | 2025 | [79] |
| DP-CTNet | 86.32 | 76.36 | – | 2025 | [87] |
| HSLabeling + SegFormer | 87.50 | – | 86.83 | 2025 | [39] |
| GFDNet | 88.41 | 81.08 | 89.42 | 2025 | [90] |
| MSEONet | 88.68 | 74.51 | 84.17 | 2025 | [19] |
| **MeCSAFNet-large (4c)** | 91.17 | 84.16 | 91.27 | 2026 | ours |
| **MeCSAFNet-base (6c)** | 91.18 | 84.14 | 91.24 | 2026 | ours |

*Dash (–) indicates metrics not reported in the original publication.*

**Table 10**
Ablation analysis of the proposed approach on the Potsdam dataset using MeCSAFNet-tiny version. Speed refers to the average inference time per image patch. The best results are underlined.

| Bands | CBAM | CAM | SAM | ReLU | ASAU | OA (%) | mIoU (%) | mF1 (%) | T. time (h) | Speed (s) |
|---|---|---|---|---|---|---|---|---|---|---|
| RGB NIR | ✓ | ✗ | ✗ | ✗ | ✓ | 90.87 | 83.56 | 90.90 | 2.16 | 0.0310 |
| | ✗ | ✓ | ✗ | ✗ | ✓ | 89.74 | 81.73 | 89.81 | 3.02 | 0.0311 |
| | ✗ | ✗ | ✓ | ✗ | ✓ | 89.79 | 81.69 | 89.78 | 2.93 | 0.0304 |
| | ✓ | ✗ | ✗ | ✓ | ✗ | 88.98 | 80.61 | 89.13 | 2.73 | 0.0303 |
| | ✗ | ✓ | ✗ | ✓ | ✗ | 89.18 | 80.85 | 89.28 | 2.69 | 0.0292 |
| | ✗ | ✗ | ✓ | ✓ | ✗ | 89.33 | 81.19 | 89.48 | 2.52 | 0.0288 |
| RGB NIR NDVI NDWI | ✓ | ✗ | ✗ | ✗ | ✓ | 90.93 | 83.72 | 91.00 | 2.83 | 0.0494 |
| | ✗ | ✓ | ✗ | ✗ | ✓ | 88.91 | 80.34 | 88.95 | 2.85 | 0.0308 |
| | ✗ | ✗ | ✓ | ✗ | ✓ | 87.61 | 78.30 | 87.74 | 2.93 | 0.0305 |
| | ✓ | ✗ | ✗ | ✓ | ✗ | 88.54 | 79.82 | 88.67 | 2.79 | 0.0300 |
| | ✗ | ✓ | ✗ | ✓ | ✗ | 89.01 | 80.43 | 89.04 | 2.70 | 0.0394 |
| | ✗ | ✗ | ✓ | ✓ | ✗ | 89.03 | 82.00 | 88.99 | 2.69 | 0.0306 |

a decoder that maintains sharp and consistent boundaries. Moreover, unlike heavy Transformer-based pipelines, our model reaches these results with moderate training epochs and no reliance on auxiliary data modalities like DSM or pseudo-labeling.

Overall, our model demonstrates a strong balance between accuracy, adaptability, and efficiency across diverse datasets and input configurations. Notably, MeCSAFNet performs well under distinct segmentation challenges. The FBP dataset demands class-level consistency and semantic precision at large scale, while the Potsdam dataset requires fine-grained delineation of small urban objects and geometrically accurate boundaries due to its very high spatial resolution. The consistent improvements observed over both traditional baselines and recent state-of-the-art methods confirm the robustness of the proposed design across varied operational conditions, supporting its suitability for land cover segmentation tasks that require both precision and deployability.

## 5. Limitations

Although MeCSAFNet has shown strong performance across different datasets and input configurations, some limitations should be acknowledged. All experiments were conducted using a dual-GPU setup with 40 GB of memory per unit, which enabled efficient training even for the largest model variants. In more resource-constrained environments, training time and memory requirements could be significantly higher, potentially limiting the accessibility or scalability of the most complex versions. Also, as with many segmentation approaches, performance may vary when dealing with highly complex scenes or different spectral resolutions, which remain challenging scenarios in remote sensing.

Additionally, the results indicate that the most complex variants, such as MeCSAFNet-large, may require longer training to fully converge due to their large number of parameters (more than double compared to the base variant), particularly when processing inputs with extended spectral channels. While the reported configurations achieved competitive performance within the predefined training setup, the convergence behavior of these larger models suggests that longer training schedules might be required in other environments to match or exceed the reported results.

Finally, the proposed 6-channel configuration incorporates NDVI and NDWI as representative hand-crafted spectral indices to enrich the non-visible input stream. While these indices are widely adopted and well-established in the literature for vegetation and water-related discrimination, they do not exhaust the set of possible spectral indices that could be considered. Other indices targeting additional land-cover

**Table 11**
Percentage improvement of our best models compared to baselines and other state-of-the-art methods on each dataset employed.

| Dataset | Reference model | ΔOA (%) | ΔmIoU (%) | ΔmF1 (%) |
|---|---|---|---|---|
| **MeCSAFNet-base (6c)** | | | | |
| FPB | U-Net (4c) | +5.09 | +19.21 | +13.01 |
| | U-Net (6c) | +3.10 | +14.72 | +10.17 |
| | DeepLabV3+ (4c) | +4.72 | +14.87 | +9.46 |
| | DeepLabV3+ (6c) | +2.92 | +9.98 | +7.14 |
| | Segformer (4c) | +4.66 | +19.62 | +13.26 |
| | Segformer (6c) | +4.43 | +14.74 | +9.53 |
| | GeoAgent + DeepLabv3+ [45] | – | +16.93 | +30.58 |
| | Mix Transformer [5] | +17.23 | +37.91 | +35.40 |
| | ConvNeXt+Transformer [35] | +4.18 | – | – |
| | U-Net + SK-ResNeXt50 [59] | +13.94 | +33.82 | – |
| | CSNet [29] | +1.43 | +3.54 | – |
| **MeCSAFNet-large (4c)** | | | | |
| Potsdam | U-Net (4c) | +3.97 | +6.72 | +3.68 |
| | U-Net (6c) | +3.29 | +6.19 | +2.92 |
| | DeepLabV3+ (4c) | +3.94 | +6.48 | +3.58 |
| | DeepLabV3+ (6c) | +3.38 | +5.85 | +3.14 |
| | Segformer (4c) | +2.58 | +9.11 | +4.93 |
| | Segformer (6c) | +2.71 | +4.80 | +2.60 |
| | DSPCANet [37] | +1.15 | +8.37 | +4.68 |
| | HSLabeling + SegFormer [39] | +4.19 | – | +5.12 |
| | CMPF-U-Net [32] | +2.53 | +4.90 | |
| | MSEONet [19] | +2.81 | +12.95 | +8.44 |
| | GFDNet [90] | +3.12 | +3.80 | +2.07 |

*Dash (–) indicates metrics not available in the original publication to compute improvement delta.*

characteristics (e.g., built-up or impervious surfaces) may also be explored. However, a systematic analysis of alternative or additional indices was beyond the scope of this work.

## 6. Conclusions and future work

This work proposes MeCSAFNet, an architecture for land cover segmentation in multispectral remote sensing imagery. The model follows a multi-branch encoder-decoder structure, where visible and non-visible channels are processed independently using ConvNeXt encoders to extract hierarchical spatial-semantic features. Each branch includes its own decoder to reconstruct spatial detail, while an additional fusion module integrates intermediate features across stages, enabling the combination of fine-grained spatial cues with high-level spectral representations. Furthermore, the decoding fusion stage is enhanced with the CBAM attention mechanism, and the use of the ASAU activation function contributes to smoother and more stable optimization. The model is designed to handle both 4-channel (RGB-NIR) and 6-channel (RGB, NIR, NDVI, NDWI) input configurations, adapting to different levels of spectral richness.

Results on the Potsdam and Five-Billion-Pixels datasets demonstrate that MeCSAFNet delivers consistent and substantial improvements over other methods. We highlight that MeCSAFNet-base (6c) improves mIoU by up to +19.62% over SegFormer (4c) and by +9.98% over DeepLabV3+ (6c) on the FPB dataset, while MeCSAFNet-large (4c) achieves a +6.19% gain in mIoU over U-Net (6c), and by +4.80 over SegFormer (6c) on Potsdam. Moreover, the availability of multiple model variants allows adaptation to different resource constraints, as the lighter versions of MeCSAFNet maintain competitive performance while reducing training complexity and memory requirements. Beyond these baselines, the proposed model also outperforms several recent state-of-the-art approaches, confirming its robustness across both traditional and modern segmentation pipelines. In addition, the inference cost of all variants remains moderate, making the architecture suitable for deployment in operational environments.

Future work could explore a broader range of experiments to evaluate the generalizability of MeCSAFNet across different datasets, including scenarios with varying spatial resolutions, sensor modalities, spectral indices, and channel configurations. This would help assess the robustness of the architecture under more diverse and challenging conditions. Additionally, there is room for further refinement of the model's design. Potential directions include replacing the current encoder with alternative feature extractors, exploring hybrid architectures (such as CNN-Transformer), or reconfiguring the decoding and fusion blocks to improve boundary accuracy and spectral integration. Finally, exploring the model in other spectral domains, such as hyperspectral imagery, could be an interesting avenue for future research.

## CRediT authorship contribution statement

**Leo Thomas Ramos:** Writing – review & editing, Writing – original draft, Methodology, Investigation, Formal analysis, Data curation, Conceptualization. **Angel D. Sappa:** Writing – review & editing, Supervision, Project administration, Methodology, Conceptualization.

## Funding

This material is based upon work supported by the Air Force Office of Scientific Research under award number FA9550-24-1-0206; and partially supported by the Grant PID2021-128945NB-I00 funded by MICIU/AEI/ 10.13039/501100011033 and by ERDF/EU and Grant PID2024-162815NB-I00 funded by MICIU/AEI/ 10.13039/501100011033 and by ERDF/EU. The authors acknowledge the support of the Generalitat de Catalunya CERCA Program to CVC's general activities..

## Declaration of competing interest

The authors declare that they have no known competing financial interests or personal relationships that could have appeared to influence the work reported in this paper.

## Data availability

Datasets used in this work can be found at https://www.isprs.org/resources/datasets/benchmarks/UrbanSemLab/2d-sem-label-potsdam.aspx (Potsdam), and at https://x-ytong.github.io/project/Five-Billion-Pixels.html (Five-Billion-Pixels).

## Author biography

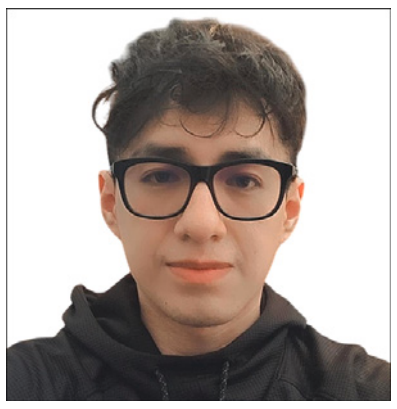

**Leo Thomas Ramos** received the degree magna cum laude in Information Technology Engineering with a focus on Artificial Intelligence from Yachay Tech University (Ecuador), in 2023. He is currently pursuing the Ph.D. degree in Computer Science with the Computer Vision Center, Universitat Autònoma de Barcelona (Spain). He held the position of Research Associate at the University of Texas at Arlington (USA), in which he supervised undergraduate and master's students in scientific research. In parallel, he serves as a Research Engineer at Kauel Inc. (Silicon Valley, USA), where he develops computer vision systems for the energy industry. His research interests include deep learning, remote sensing, Earth observation, space exploration, planetary science, multimodal and multispectral data analysis, and medical imaging. He has collaborated with multidisciplinary teams across Europe, the Middle East, Asia, and the Americas, including renowned researchers from Dubai Police, the Environment Agency Abu Dhabi, Halliburton, Saudi Aramco, Honeywell, the Air Force Research Laboratory, and NASA Jet Propulsion Laboratory.

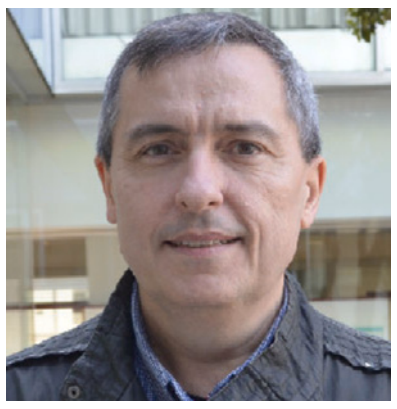

**Angel D. Sappa** received the degree in Electromechanical Engineering from the National University of La Pampa, Argentina, in 1995, and the Ph.D. in Industrial Engineering from the Polytechnic University of Catalonia, Spain, in 1999. He is currently a Senior Scientist at the Computer Vision Center, Universitat Autònoma de Barcelona, Spain, and a Full-Time Professor at ESPOL Polytechnic University, Ecuador, where he leads the computer vision team at the CIDIS Research Center. He serves as Associate Editor for Pattern Recognition, and as Guest Editor for Robotics and Autonomous Systems and the Journal of Computational Science. He has over 25 years of experience in computer vision and artificial intelligence, with more than 300 publications in top-ranked journals and leading international conferences. In 2023 and 2024, he was included in Stanford University's top 2% most-cited researchers in the subfield of artificial intelligence (career-long ranking).